\pdfoutput=1

\documentclass[11pt]{article}

\usepackage[]{acl}
\usepackage{graphicx}
\usepackage{subfigure}
\usepackage{tabularx} 

\usepackage{multirow}   
\usepackage{xcolor}     
\usepackage{colortbl}
\usepackage{amsmath,amsfonts}
\usepackage{algorithmic}
\usepackage{adjustbox}
\usepackage{algorithm}
\usepackage{array}
\usepackage{subfigure}
\usepackage{subcaption}
\usepackage{textcomp}
\usepackage{stfloats}
\usepackage{url}
\usepackage{verbatim}
\usepackage{graphicx}
\usepackage{booktabs}
\usepackage{graphicx}%
\usepackage{float}
\usepackage{caption}
\usepackage{times}
\usepackage{latexsym}
\usepackage{natbib}
\usepackage{balance}
\usepackage[normalem]{ulem}

\usepackage{arydshln}

\usepackage[T1]{fontenc}

\usepackage[utf8]{inputenc}

\usepackage{microtype}

\usepackage{inconsolata}

\usepackage{graphicx}

\def\method{ConsJudge}

\title{Judge as A Judge: Improving the Evaluation of Retrieval-Augmented Generation through the Judge-Consistency of Large Language Models}

\author{Shuliang Liu$^{1}$\thanks{ \ \ indicates equal contribution.}, Xinze Li$^{1}$\footnotemark[1], Zhenghao Liu$^{1}$\thanks{ \ \ indicates corresponding author.}, Yukun Yan$^{2}$,\\ \textbf{Cheng Yang$^{3}$, Zheni Zeng$^{2}$, Zhiyuan Liu$^{2}$, Maosong Sun$^{2}$,  Ge Yu$^{1}$} \\ 
$^1$Department of Computer Science and Technology, Northeastern University, China \\
$^2$Department of Computer Science and Technology, Institute for AI, Tsinghua University, China \\
Beijing National Research Center for Information Science and Technology, China \\
$^3$School of Computer Science, Beijing University of Posts and Telecommunications\\
}

\begin{document}
\maketitle

\begin{abstract}
Retrieval-Augmented Generation (RAG) has proven its effectiveness in alleviating hallucinations for Large Language Models (LLMs). However, existing automated evaluation metrics cannot fairly evaluate the outputs generated by RAG models during training and evaluation. LLM-based judgment models provide the potential to produce high-quality judgments, but they are highly sensitive to evaluation prompts, leading to inconsistencies when judging the output of RAG models. 
This paper introduces the Judge-Consistency (\method{}) method, which aims to enhance LLMs to generate more accurate evaluations for RAG models. Specifically, \method{} prompts LLMs to generate different judgments based on various combinations of judgment dimensions, utilize the judge-consistency to evaluate these judgments and select the accepted and rejected judgments for DPO training. 
Our experiments show that \method{} can effectively provide more accurate judgments for optimizing RAG models across various RAG models and datasets. Further analysis reveals that judgments generated by \method{} have a high agreement with the superior LLM. All codes are available at \url{https://github.com/OpenBMB/ConsJudge}.
\end{abstract}


\section{Introduction}
Retrieval-Augmented Generation (RAG)~\cite{REALM2020Guu,Retrieval2020Lewis,asai2023self,shi2023replug} has proven effective in mitigating hallucinations~\cite{elazar2021measuring,ji2023survey,shuster2021retrieval,huang2023survey} in Large Language Models (LLMs). RAG retrieves relevant knowledge from the knowledge base and incorporates the external information as the input context~\cite{Ram2023Incontextlearning}, benefiting various knowledge-intensive tasks~\cite{trivedi2023interleaving,izacard2022few,he2021efficient}. Existing studies typically employ automated evaluation metrics to assess the outputs of RAG systems during both training~\cite{rag-ddr2024Li} and evaluation~\cite{Radit2023Lin,Retrieval2023Gao} phases. These metrics primarily focus on string-level exact matching, which is less effective for determining whether the LLM-generated responses align with the ground truth~\cite{EnablingLargeLanguageModelstoGenerateTextwithCitations2023GaoTianyu,saad2024ares}, posing a challenge for RAG systems.

\begin{figure}[t]
    \centering
    \includegraphics[width=\columnwidth]{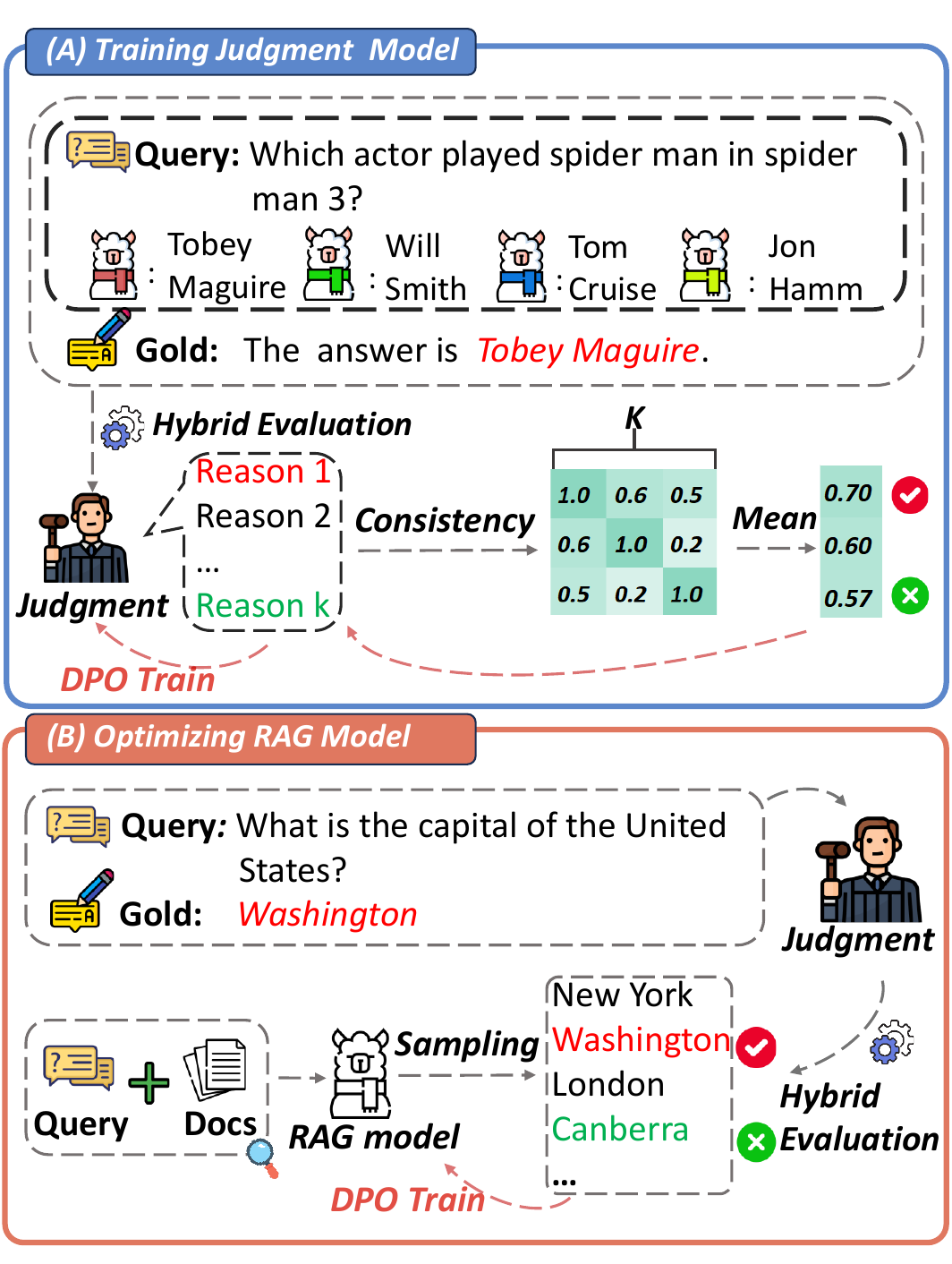}
    \caption{The Framework of \method{}. It enhances the judgment capabilities of LLMs and benefits the training process of RAG models.}
    \label{fig:example}
\end{figure}


To enable more accurate evaluation of RAG models, some works incorporate LLMs as judgment models to assess the quality of generated responses~\cite{saad2024ares,friel2024ragbench,adlakha2023evaluating}, relying on their human-equivalent evaluation capabilities~\cite{chiang2023can,zheng2023judging,sottana2023evaluation}. These judgment models introduce specific evaluation dimensions, such as hallucination and comprehensiveness, to prompt LLMs to verify whether the generated responses align with the facts in the retrieved documents and whether all relevant information has been properly extracted and integrated~\cite{jin2024rag,jacovi2025facts,Rageval2024Zhu}. However, LLM-based judgment models are highly sensitive to the design of evaluation prompts~\cite{zhou2023survival,liu2024aligning}, which may lead to inconsistencies in judgments when different evaluation dimensions are employed.

In this paper, we introduce Judge-Consistency (\method{}), a method that enhances LLM-based judgment models to generate more accurate evaluations for RAG models in a self-improvement framework. \method{} incorporates specialized evaluation dimensions and employs a multiple-choice selection strategy for evaluation modeling. Additionally, \method{} leverages the Direct Preference Optimization (DPO) method~\cite{DPO2023Rafailov} to enhance the judgment capabilities of LLMs, while also implementing a ``judge as a judge'' mechanism during choosing preference pairs of judgments. Specifically, \method{} encourages LLMs to generate judgment results based on various combinations of judgment dimensions, evaluates the quality of these judgments using judge-consistency, and selects the chosen and rejected judgments for DPO training. This approach enhances the performance of the LLM-based judgment model without necessitating distillation from more powerful LLMs~\cite{zhang2025rag}.

Our experiments demonstrate the effectiveness of the \method{} method as a reward model for optimizing the RAG model, resulting in significant improvements over vanilla LLMs. Further analysis shows that \method{} helps LLMs select appropriate evaluation dimensions for assessing response quality by incorporating the agreements and consistency across different evaluation dimensions during optimization. Additionally, compared to other baselines, \method{} exhibits higher judgment consistency with the superior LLM, GLM-4-plus~\cite{du2022glm}, across various RAG evaluation datasets, further highlighting its effectiveness in optimizing LLMs to produce more accurate judgments on diverse RAG tasks.

\section{Related Work}
Retrieval-Augmented Generation (RAG) has proven its effectiveness in various tasks, such as open-domain question answering~\cite{trivedi2023interleaving}, dialogue systems~\cite{cai2019skeleton}, and code generation~\cite{li2025building}. By integrating external knowledge into the input context~\cite{Ram2023Incontextlearning}, Large Language Models (LLMs) can alleviate hallucination issues and produce more accurate responses~\cite{RALM2024Asai}. To evaluate the performance of LLM responses, most existing RAG models rely on automatic metrics like Exact-Match during both evaluation~\cite{Radit2023Lin,Retrieval2023Gao} and training~\cite{rag-ddr2024Li}. However, these metrics may fail to offer a fair assessment when the LLM-generated responses are lengthy or semantically similar to the ground truth but not character-matched~\cite{EnablingLargeLanguageModelstoGenerateTextwithCitations2023GaoTianyu}.

Recently, LLMs, such as ChatGPT~\cite{openai2023gpt}, have demonstrated human-equivalent performance~\cite{chiang2023can,zheng2023judging,sottana2023evaluation} and are now commonly used as judgment models to assess generated responses in various tasks~\cite{chen2023huatuogpt,chen2023teaching,an2023eval,chanchateval}. These studies typically prompt LLMs to evaluate generated responses based on specific evaluation dimensions~\cite{chiang2023closer}. However, using APIs of these closed-source models incurs significant costs and reduces reproducibility, as the models may evolve behind the API~\cite{SurveyonLLM-as-a-Judge2024Gu}. To address this, some researchers are turning to open-source LLMs as alternatives, using judgments from closed-source LLMs to fine-tune open-source models and improve their evaluation capabilities~\cite{wangpandalm,zheng2023judging}.

Moreover, RAG models also employ LLMs as judges to assess the quality of generation, focusing on the relevance and faithfulness of the responses~\cite{saad2024ares,friel2024ragbench,adlakha2023evaluating}. These models typically design prompts to instruct LLMs to determine whether the generated response aligns with the facts in the retrieved document and whether all relevant information has been fully extracted and integrated~\cite{jin2024rag,jacovi2025facts,Rageval2024Zhu}. However, LLM-based evaluations are highly sensitive to prompt designs~\cite{zhou2023survival,liu2024aligning}, making the judgments inconsistent when using different dimensions for evaluating RAG responses. To mitigate this issue, \method{} introduces a judge-consistency approach to self-improve the judgment performance of LLMs in RAG systems, avoiding distillation from larger-scale LLMs~\cite{zhang2025rag}.

\section{Methodology}
In this section, we introduce the Judge-Consistency (\method{}) method, which optimizes Large Language Models (LLMs) as judgment models to evaluate the effectiveness of the Retrieval-Augmented Generation (RAG) system. First, we discuss the evaluation methods used in \method{} (Sec.\ref{sec:3.1}). Next, we present how the \method{} method is utilized to optimize the judgment model, improving its judgment accuracy (Sec.\ref{sec:3.2}). Finally, we apply the \method{} method as the reward model to optimize the RAG system (Sec.~\ref{sec:3.3}). \begin{figure*}
\centering
\includegraphics[width=\linewidth]{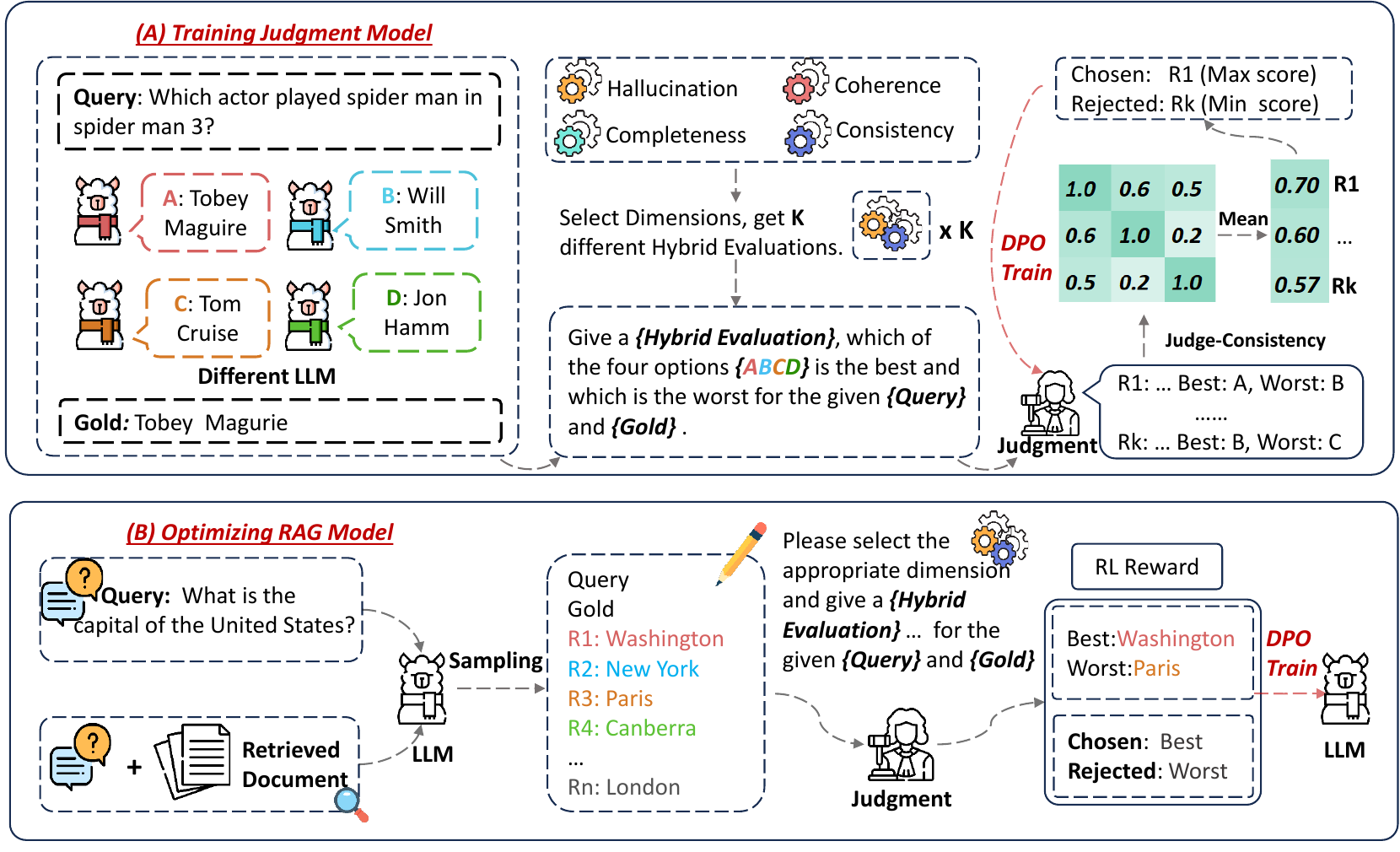}
\caption{The Framework of Our \method{} Method.} 
\label{fig:model}
\end{figure*}

\subsection{Preliminary of RAG Evaluation Methods} \label{sec:3.1}
Recent work, such as LLM-as-a-Judge~\cite{zheng2023judging}, typically regards LLMs as judgment models for rating responses in various NLP tasks. These methods use specially designed prompts to employ LLMs in evaluating generated responses across different dimensions, such as hallucination and accuracy~\cite{li2024llms}. In this section, we describe the evaluation dimensions and modeling methods used in \method{}.

\textbf{Evaluation Dimensions.} Some studies~\cite{Rageval2024Zhu,zhang2025rag} evaluate the generation quality of RAG models based on multiple criteria. \method{} follows these approaches by designing different evaluation prompts for four key dimensions: hallucination, completeness, coherence, and semantic consistency.

\textit{Hallucination.} Hallucination refers to the inclusion of information in the response that contradicts the ground truth. This dimension aims to detect whether the generated responses contain factual errors due to hallucinations~\cite{xu2024hallucination}.

\textit{Completeness.} Completeness evaluates whether the generated responses contain as much relevant information as possible from the ground truth. This dimension primarily focuses on identifying whether the responses omit some key points from the ground truth answers.

\textit{Coherence.} Coherence evaluates whether the responses are logically consistent and whether the language flows fluently between sentences. This dimension is primarily concerned with ensuring that the responses are both coherent and fluent.

\textit{Semantic Consistency.} Semantic consistency checks whether the generated response is semantically aligned with the ground truth answer, rather than simply matching it lexically. This dimension helps avoid misjudging responses that differ from the ground truth in terms of tokens but share the same meaning with ground truth answers.

\textbf{Evaluation Modeling.} For evaluation modeling, our \method{} method uses a multiple-choice selection approach~\cite{SurveyonLLM-as-a-Judge2024Gu}. In this method, LLMs evaluate all candidate responses and choose the best or the worst based on different evaluation dimensions. This facilitates RAG training using the judgment model (Sec.~\ref{sec:3.3}).

Existing methods typically rely on Pointwise Evaluation or Listwise Comparison for candidate response evaluation. Pointwise Evaluation directly prompts LLMs to score each candidate based on predefined evaluation dimensions.
However, this method fails to capture the differences between responses, leading to evaluation bias~\cite{kim2023prometheus,wang2023learning}. 
In contrast, Listwise Comparison prompts LLMs to evaluate an entire list of candidate responses and rank them~\cite{niu2024judgerank,yan2024consolidating}, allowing for a more comprehensive evaluation~\cite{li2024llms}. The \method{} model adopts a multiple-choice selection method, which performs Listwise Comparison to evaluate all candidate responses.



\subsection{Training Evaluation Models through Judge-Consistency} \label{sec:3.2}
Although LLMs have demonstrated effectiveness in evaluating their own responses, judgment models may still suffer from issues such as Position Bias, Verbosity Bias, and Evaluation Metric Bias, which can compromise the quality of judgments~\cite{zheng2023judging,chen2024humans,li2024llms}. To address these challenges, we propose the Judge-Consistency method to optimize the judgment model $\mathcal{M}$ based on the consistency of judgments across different evaluation dimensions. This process self-improves the judgment model by selecting more suitable evaluation dimensions, ultimately allowing for more precise assessments.

\textbf{Evaluation of LLM Responses.} To optimize the judgment model, we begin by selecting $m$ LLMs and sampling responses from each using different temperatures. Next, we randomly select one response $y$ from each LLM, forming a response set $Y=\{y_1, \dots, y_m\}$. We then combine the four evaluation dimensions introduced in Sec.~\ref{sec:3.1}, generating $k$ distinct hybrid evaluation aspects:
\begin{equation}\small
\mathcal{I} = \{I_1,...,I_k\},
\end{equation}
where $I_i$ represents a hybrid evaluation aspect, which could be a single evaluation dimension or a combination of multiple dimensions. For each evaluation aspect $I_i \in \mathcal{I}$, we create an evaluation prompt, yielding $k$ distinct prompts:
\begin{equation}\small
\mathcal{P} = \{P^1,...,P^k\},
\end{equation}
where $P^i$ is the $i$-th evaluation prompt. The judgment model $\mathcal{M}$ then generates a judgment result $r_i$ for evaluating LLM-generated responses $Y$, based on the $i$-th evaluation aspect $I_i$:
\begin{equation}\small
r_i = \mathcal{M}(P^i,q,y^*, Y),
\end{equation}
where $y^*$ is the ground truth for question $q$. The judgment result $r_i$ includes both the best ($y^+$) and the worst ($y^-$) selections from the candidate responses $Y = \{y_1, \dots, y_m\}$, along with chain-of-thoughts~\cite{Chain-of-Thought2022Wei} for the judgment. By utilizing all evaluation prompts in $\mathcal{P}$, we obtain $k$ judgment results, denoted as $R = \{r_1, \dots, r_k\}$.

\textbf{Judge Consistency Evaluation.} We follow the previous work~\cite{self-improve2023Li} to introduce a ``Judge as a judge'' approach that evaluates the consistency of judgments across different prompts.

Specifically, after obtaining the judgment results $R = \{r_1, \dots, r_k\}$, we use a text embedding model $\text{Emb}(\cdot)$ to compute the similarity score between the $i$-th judgment $r_i$ and all other judgment results $R$. The average of these similarity scores provides the consistency score $S_i$ for the judgment $r_i$:
\begin{equation}\small\label{eq:score}
S_i = \frac{1}{k} \sum_{j=1}^{k} \cos (\text{Emb}(r_i),\text{Emb}(r_j)).
\end{equation}
Judgments exhibiting higher consistency scores are considered positive results, while those with lower consistency scores are interpreted as negative results, indicating potential judge bias.

\textbf{Judgment Model Optimization.} We then optimize the judgment model $\mathcal{M}$ to better select the appropriate evaluation aspect from the set $\mathcal{I}$ to make more accurate judgments.

To achieve this, we treat the judgment with the highest consistency score as a positive judgment $r^+$, and the judgment with the lowest score as the negative one $r^-$. We collect the instance $(q, y^*, r^+, r^-)$ to form the training dataset $\mathcal{T}$. The judgment model $\mathcal{M}$ then uses prompts $\mathcal{P}$ for evaluating the quality of responses $Y$ and is optimized to assign a higher probability to the positive judgment $r^+$ than to the negative judgment $r^-$. This is accomplished through the Direct Preference Optimization (DPO) method~\cite{DPO2023Rafailov}:
\begin{equation}\small\label{eq:dpo}
\begin{aligned}
 & \mathcal{L}= - \mathbb{E}_{(q, y^*, r^+, r^-) \sim \mathcal{T}} [\log \sigma(\beta \\ &\log \frac{\mathcal{M}(r^+ \mid q, y^*)}{\mathcal{M}^\text{ref}(r^+ \mid q, y^*)} - \beta \log \frac{\mathcal{M}(r^- \mid q, y^*)}{\mathcal{M}^\text{ref} (r^- \mid q, y^*)})],
\end{aligned}
\end{equation}
where $r^+, r^- \in R$, and $\beta$ is a hyperparameter. $\mathcal{M}^\text{ref}$ denotes the reference model, which remains fixed during training. 

These judgment results in $R$ are generated using different evaluation prompts $\mathcal{P}$ that combine various evaluation dimensions. \method{} optimizes the judgment model $\mathcal{M}$ to reproduce the positive judgment $r^+$ that shares the most consistency score with others, making the judgment model $\mathcal{M}$ select more appropriate dimensions to evaluate the response quality of LLMs.

\subsection{Applying Judgment Models to Optimize Retrieval-Augmented Generation Systems}
\label{sec:3.3}
To evaluate the effectiveness of judgment model $\mathcal{M}$, we use it as the reward model and apply DPO to optimize the RAG system~\cite{rag-ddr2024Li}.

For a given query $q$, the current RAG system typically utilizes a dense retriever model to retrieve the Top-$n$ relevant documents $\mathcal{D} = \{d_1, \dots, d_n\}$ from the external knowledge bases. The generation model ($\text{Gen}$) then samples outputs $\Tilde{y}$, either with or without the retrieved documents $\mathcal{D}$:
\begin{equation}\small
\begin{aligned}
    \Tilde{y} &\sim \text{Gen} (\mathcal{D} \oplus q), \\
    \Tilde{y} &\sim \text{Gen} (q),
\end{aligned}
\end{equation}
where $\oplus$ is the concatenation operation. Then, we collect all sampled responses $\Tilde{y}$ in the set $\Tilde{Y}$ and utilize the judgment model $\mathcal{M}$ to generate the judgment result $r_\text{all}$ based on all evaluation dimensions:
\begin{equation}\small
r_\text{all} = \mathcal{M}(P_\text{all}, q,y^*,\Tilde{Y}),
\end{equation}
where $P_\text{all}$ indicates the evaluation prompt that involves all evaluation dimensions. Based on $r_\text{all}$, we select the best response $\Tilde{y}^+$ and worst response $\Tilde{y}^-$ from $\Tilde{Y}$, respectively. This allows us to construct a preference dataset $(q, \mathcal{P}, \Tilde{y}^+, \Tilde{y}^-)$ to optimize the generation model ($\text{Gen}$) via DPO training. If the optimized generation model ($\text{Gen}$) demonstrates improved performance on the downstream RAG tasks, this indicates that the judgment model $\mathcal{M}$ provides more precise judgments, effectively serving as the reward for optimizing the RAG system.

\section{Experimental Methodology}
This section describes the datasets, evaluation metrics, baselines, and implementation details used in our experiments. More implementation details are shown in Appendix~\ref{app:dataset details}.

\begin{table}[t]
\centering
\small
\begin{tabular}{l|l|r}
\hline
\textbf{Task} & \textbf{Dataset} & \textbf{Total} \\
\hline
\multirow{3}{*}{Open-Domain QA} 
& NQ~\shortcite{nq2019Kwiatkowski} & 2,837 \\
& TriviaQA~\shortcite{triviaqa2017Joshi} & 5,359 \\
& MARCOQA~\shortcite{bajaj2016ms} & 1,000 \\
\hline
Factoid QA & ASQA~\shortcite{ASQA2022Stelmakh} & 958 \\
\hline
Multi-Hop QA & HotpotQA~\shortcite{hotpotqa2018Yang} & 5,600 \\
\hline
Dialogue & WoW~\shortcite{wow2019Dinan} & 1,000 \\
\hline
\end{tabular}
\caption{Data Statistics of the RAG Evaluation Datasets.}
\label{table1:testdataset}  
\end{table}

\textbf{Datasets.} We describe the datasets used for training \method{} and RAG training and evaluation.

\textit{\method{} Training.} For training \method{}, we collect 11 knowledge-intensive tasks from previous works~\cite{Chung2022flan_t5,izacard2022few} to collect 73,831 instances and 3,886 instances to construct both training and development sets.


\textit{RAG Training \& Evaluation.} To retrieve documents for constructing the RAG datasets, we use BGE-large~\citep{bge_embedding} with the MS MARCO V2.1~\citep{bajaj2016ms} corpus. During RAG training, we collect seven datasets from ~\citet{rag-ddr2024Li} and randomly sample 20,805 samples for the training set and 1,400 samples for the development set. For RAG evaluation, we select knowledge-intensive tasks from prior work~\citep{rag-ddr2024Li, xu2024unsupervised}, including open-domain QA tasks (NQ~\citep{nq2019Kwiatkowski}, TriviaQA~\citep{triviaqa2017Joshi}, MARCO QA~\citep{bajaj2016ms}), multi-hop QA (HotpotQA~\cite{hotpotqa2018Yang}), factoid QA (ASQA~\cite{ASQA2022Stelmakh}), and dialogue tasks (WoW~\cite{wow2019Dinan}). The data statistics are shown in Table~\ref{table1:testdataset}.


\textbf{Evaluation Metrics.} 
For tasks with longer outputs, automated evaluation metrics, such as ROUGE, cannot evaluate the quality of outputs fairly, which has been proven by previous work~\cite{EnablingLargeLanguageModelstoGenerateTextwithCitations2023GaoTianyu,llmeval2024zhang}. Thus, we adopt the LLM-as-a-Judge method~\citep{llmeval2024zhang}, which employs GLM-4-plus\footnote{\url{https://open.bigmodel.cn/}} for evaluation MARCO QA and WoW. Besides, we use StringEM as the evaluation metric for the ASQA dataset. For other evaluation tasks, we evaluate performance using Accuracy. The prompt using GLM-4-plus to evaluate is shown in Appendix~\ref{sec:prompt details}.

\textbf{Baselines.} In our experiments, we compare \method{} with three judgment models, including the Raw Metric model and two LLM-based judgment models. For the Raw Metric model, we utilize the automatic evaluation metrics, ROUGE-L and Accuracy, as the judgment model to optimize the RAG system. Specifically, the Raw Metric model uses ROUGE-L for MARCO QA, Yahoo!QA and WikiQA datasets, and also uses Accuracy for the remaining datasets, which is the same as previous work~\cite{rag-ddr2024Li}. Additionally, we employ two LLM-based judgment model baselines: Vanilla LLM and SFT. The Vanilla LLM method directly uses the LLM as the judgment model and then leverages the evaluation prompts to ask them to produce the judgments. The SFT method further fine-tunes LLMs based on judgment results generated by a superior LLM, GLM-4-plus, which has been used in previous work~\cite{zhang2025rag} to improve the judgment performance of LLMs.

\begin{table*}[ht]
\centering
\small
\resizebox{\textwidth}{!}{
\begin{tabular}{l|l|ccccccc}
\hline

 \multirow{2}{*}{\textbf{Generator}} & \multirow{2}{*}{\textbf{Reward}} & \textbf{NQ} & \textbf{HotpotQA} & \textbf{TriviaQA}  & \textbf{ASQA} & \textbf{MARCOQA} & \textbf{WoW} & \multirow{2}{*}{\textbf{Avg.}} \\ 
& & (acc) & (acc) & (acc) & (str-em) & (llm) & (llm) & \\ 
\hline

\multirow{7}{*}{\shortstack{\textbf{MiniCPM}\\(2.4B)}} & Raw Metric & 46.14	& 30.09	& 80.03 & 33.44 & 84.75 & 85.48 & 59.99\\  \cdashline{2-9}
& Llama3-8B& \textbf{47.23} & 29.64 & 80.40 & \textbf{35.77}	& 	86.00 & 85.98 &60.84 \\
& w/ SFT & 47.02 & 29.55 & 80.33	& 34.60	& \textbf{86.35}	& 85.98 &60.64 \\
&w/ \method{}  & 47.02 & \textbf{30.45} & \textbf{80.80}	& 35.68	& 86.16	& \textbf{86.13} & \textbf{61.04}
\\\cdashline{2-9}

& Qwen2.5-14B & 47.30 &  28.96 & 80.03	& 34.95	& 85.59	& 87.49 &60.72  \\
& w/ SFT & 47.02 & 27.64 & 79.80	& 33.51	& 	\textbf{85.90} & \textbf{87.51} & 60.23\\
& w/ \method{} & \textbf{48.01} & \textbf{30.84} & \textbf{80.69} & 	\textbf{36.45} & 85.73	& 87.21 & \textbf{61.49} \\

\hline

\multirow{4}{*}{\shortstack{\textbf{Llama3}\\(8B)}} & Raw Metric &48.96 & 36.95& 86.83	& 41.55	& 	84.80& 82.66 &63.63 \\ \cdashline{2-9}
& Llama3-8B & 46.63 &  32.84 & 85.13 & 40.69	& 	88.15 & 88.31 &63.63 \\
& w/ SFT & 47.16 & 35.95  & 86.10	& 40.37	& 87.46	& 87.97	 & 64.20 \\
&w/ \method{}  & \textbf{48.78} & \textbf{37.54} & \textbf{88.26}& \textbf{42.44}	& \textbf{88.25}	& \textbf{88.87}	 & \textbf{65.69} \\\hline
\end{tabular}
}
\caption{Overall Performance of RAG Models Optimized Using Different Judgment Models. The \textbf{best} result is highlighted. We implement the generators of RAG models by using MiniCPM-2.4B and Llama3-8B.}
\label{table1:overall}
\end{table*}

\textbf{Implementation Details.} In our experiments, we leverage LoRA~\citep{hulora} for efficient training LLMs. We set max\_epoch to 3, learning rate to 5e-5, and the warmup ratio to 0.1. For the generation model in the RAG system, we employ the MiniCPM-2.4B~\cite{minicpm-2b2024Hu} and Llama3-8B-Instruct~\cite{touvron2023llama} as the generation models. For the judgment model, we use Llama3-8B-Instruct and Qwen2.5-14B-Instruct~\cite{qwen2.5-14b2023Bai} as the backbone models. While training the judgment model, we synthesize 8 different hybrid evaluation aspects for generating the judgment results. We use MiniCPM-Embedding\footnote{\url{https://huggingface.co/openbmb/MiniCPM-Embedding}} to assess the similarity among judgments.



\section{Evaluation Result}
In this section, we first show the performance of \method{} by regarding it as a reward model to optimize the RAG model. Then we conduct the ablation studies to show the effectiveness of different modules in \method{}. Subsequently, we evaluate the judgment quality generated by \method{} and explore the consistency of judgments across different evaluation dimensions. Finally, case studies are conducted.

\subsection{Overall Performance}
\label{sec:5.1}
In this experiment, we treat these judgment models as reward models for optimizing the RAG model~\cite{rag-ddr2024Li} and evaluate their effectiveness by examining the RAG performance.

As shown in Table~\ref{table1:overall}, we compare \method{} with three judgment models, including Raw Metric, vanilla LLMs, and SFT. Specifically, the Raw Metric model uses an automatic metric as the reward model, while vanilla LLMs and SFT rely on LLMs as reward models. In our experiments, we use Llama3-8B and Qwen2.5-14B to implement the judgment models, and both MiniCPM-2.4B and Llama3-8B to build the RAG model.

Overall, \method{} outperforms all baseline models across different RAG models, demonstrating its effectiveness in judging sampled responses of RAG models. Compared to the Raw Metric model, LLM-based judgment models achieve better optimization performance for the RAG model, even when the Raw Metric model directly optimizes RAG models to align with final evaluation metrics. It shows that LLMs have the ability to produce high-quality judgments, benefiting the training and evaluation processes of RAG systems.
In comparison to vanilla LLMs, the SFT model uses labels produced by GLM-4-plus for training, while \method{} introduces a judge-consistency based optimization method that trains LLMs without relying on additional training signals. \method{} not only outperforms the SFT model but also achieves more improvements over the Raw Metric model, highlighting the effectiveness of our judge-as-a-judge mechanism in enhancing LLMs through a self-improvement approach. Notably, the evaluation results demonstrate that the advantages of \method{} can be extended to various RAG scenarios and judgment models of different scales.

\subsection{Ablation Study}
\label{sec:5.2}
This experiment conducts ablation studies to investigate the contribution of different modules in \method{}.

As shown in Table~\ref{table1:ablation}, we evaluate the judgment performance of three variants of the \method{} model by reranking the sampled responses from vanilla RAG models and calculating the accuracy of the top-1 ranked responses. The three models compared in the experiments are: \method{} w/o Consistency, \method{} w/o Query, and \method{} w/o Ground Truth. Specifically, \method{} w/o Consistency randomly selects the chosen and rejected responses for DPO training. Both \method{} w/o Query and \method{} w/o Ground Truth models are evaluated by removing the query and ground truth from the input prompts, respectively.

\begin{table}[t]
\centering
\resizebox{\linewidth}{!}{
\begin{tabular}{lccccc}
\hline
\multirow{1}{*}{\textbf{Reward Model}} & \textbf{NQ} & \textbf{HotpotQA} & \textbf{TriviaQA} & \multicolumn{1}{c}{\multirow{1}{*}{\textbf{Avg.}}}\\

\hdashline
\multicolumn{5}{l}{\textit{Llama3-8B-Instruct}}\\
\hdashline
\method{} &  \textbf{46.70} & \textbf{30.18} & \textbf{80.13}  & \textbf{52.34}\\
w/o Consistency & 44.55 & 28.91 & 77.96 & 50.47\\
w/o Query &  44.66 & 28.86 & 75.48 & 49.67\\
w/o Ground Truth &  33.35 & 22.27 & 67.61 & 41.07\\ 

\hdashline
\multicolumn{5}{l}{\textit{Qwen2.5-14B-Instruct}}\\
\hdashline
\method{} &  \textbf{48.00} & \textbf{30.95} & \textbf{81.89} & \textbf{53.61}\\
w/o Consistency & 47.73 & 30.67 & 81.47 & 53.29\\
w/o Query  & 47.41	& 30.61 & 81.29 & 53.10\\
w/o Ground Truth & 38.03 & 25.86 & 75.78 & 46.56\\
\hline
\end{tabular}
}
\caption{Ablation Study. We use MiniCPM-2.4B to implement the RAG model.}

\label{table1:ablation}  
\end{table}

Compared to \method{} w/o Consistency, \method{} achieves a higher accuracy score. This demonstrates that the judge-consistency method enhances the ability of LLMs to select higher-quality responses, thereby benefiting the RAG training process. Furthermore, removing the query or the ground truth answer from the evaluation prompts results in a performance decrease for \method{}, although the performance gap is narrowed when using larger-scale LLMs. This suggests that both the query and ground truth help the LLM-based judgment models produce more comprehensive evaluations. Additionally, LLMs of a larger scale can leverage their parametric knowledge to assess the quality of RAG responses.

\begin{figure}[t]
    \subfigure[Judgment Consistency.] { 
    \label{fig:result:quality:agreement} 
    \includegraphics[width=0.45\linewidth]{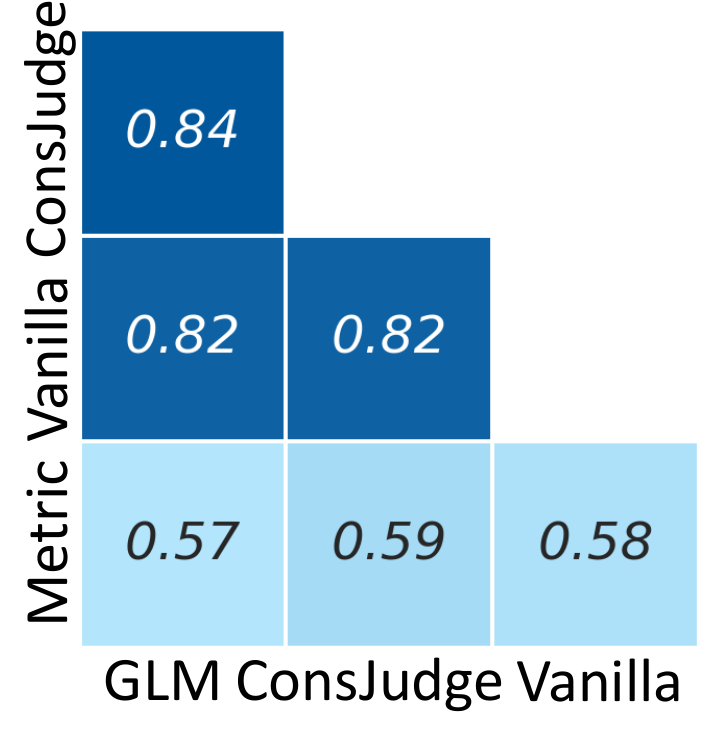}}  
    \subfigure[Agreement with GLM-4.] { 
    \label{fig:result:quality:glm} 
    \includegraphics[width=0.51\linewidth]{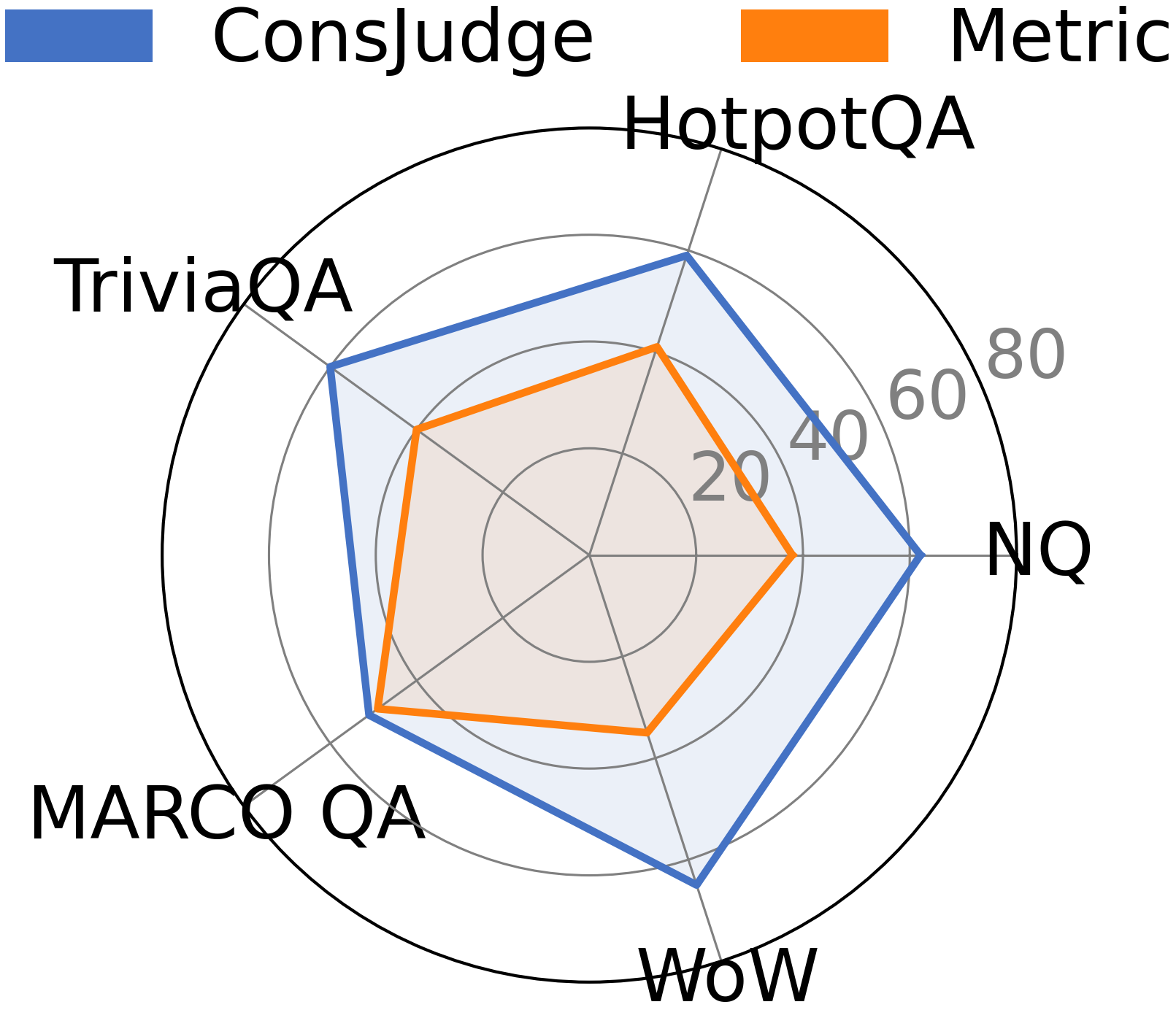}}  
    \caption{Judge Agreement Evaluation. We analyze the agreements of different judgment models (Figure~\ref{fig:result:quality:agreement}), and use GLM-4-plus to evaluate the judge quality of different models (Figure~\ref{fig:result:quality:glm}). GLM and Metric denotes the GLM-4-plus and Raw Metric models. Vanilla LLM and \method{} are implemented with Qwen2.5-14B.}
    \label{fig:result:quality}
\end{figure}

\begin{figure}[t]
    \subfigure[Llama3-8B.] { 
    \label{fig:llama} 
    \includegraphics[width=0.48\linewidth]{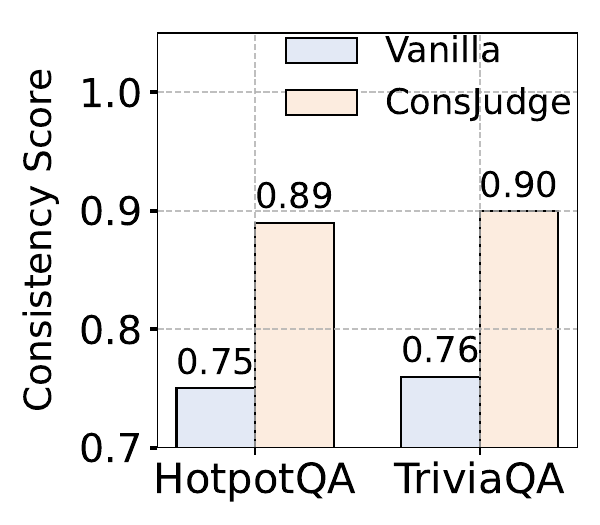}}  
    \subfigure[Qwen2.5-14B.] { 
    \label{fig:qwen} 
    \includegraphics[width=0.48\linewidth]{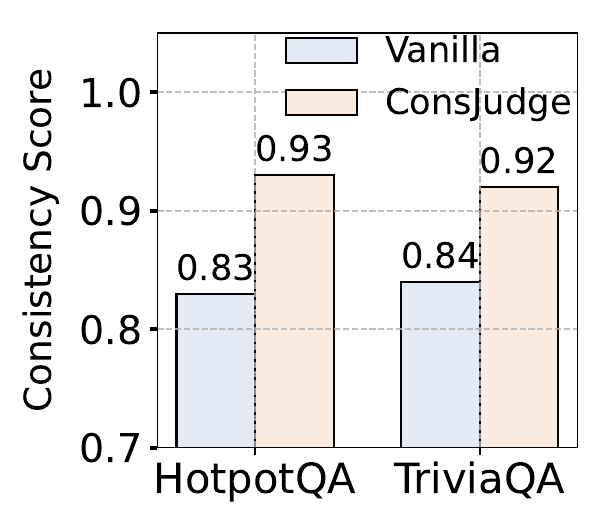}}  
    \caption{Judgment Consistency of Vanilla LLMs and \method{}. We use both vanilla LLMs and \method{} to show the judgment consistency among all hybrid evaluation aspects used to train \method{}.}
    \label{fig:consjudge_vs_vanilla_consistency}
\end{figure}

\subsection{The Judgment Quality of \method{}} \label{sec:5.3}
In this section, we first present the judge agreement across different models. Then, we analyze the consistency of judgments generated by different models based on different evaluation dimensions.

\textbf{Judge Agreement.} First, we sample 100 queries from each of the datasets--NQ, HotpotQA, TriviaQA, MARCO QA, and WoW--to construct the evaluation dataset. We then collect responses from different models and ask judgment models to evaluate these responses. As shown in Figure~\ref{fig:result:quality}, we use four judgment models: GLM-4-plus~\cite{du2022glm}, \method{}, vanilla LLM, and Raw Metrics, to select the best response for each query.

Figure~\ref{fig:result:quality:agreement} presents the judge agreement between different models. Among all judgment models, the Raw Metric shows the lowest agreement with the others, highlighting that string matching alone does not provide high-quality judgment for evaluation. Notably, \method{} not only demonstrates the highest agreement with the superior LLM, GLM-4-plus, but also conducts the agreement with GLM-4-plus when evaluating the judgments produced by vanilla LLM. This illustrates that \method{} conducts more consistent judgments with GLM-4-plus. Furthermore, we evaluate the judge agreement with GLM-4-plus of both Raw Metric and \method{} models in Figure~\ref{fig:result:quality:glm}. Across all datasets, \method{} achieves higher agreement scores with GLM-4-plus, demonstrating its effectiveness in optimizing LLMs to generate more accurate judgments in various scenarios.

\textbf{Judgment Consistency.} We next randomly sample 1,000 queries each from HotpotQA and TriviaQA to construct a dataset for evaluating judgment consistency. We then ask both vanilla LLM and \method{} to perform judgments for each query, using different hybrid evaluation aspects, and compute the consistency scores of these judgments.

As shown in Figure~\ref{fig:consjudge_vs_vanilla_consistency}, the consistency scores of judgments generated by \method{} outperform those generated by vanilla LLM across both datasets. These results indicate that \method{} achieves higher consistency with different evaluation dimensions, demonstrating its ability to comprehensively incorporate tailored evaluation dimensions to produce reliable evaluation results. Notably, the advantages of \method{} are consistent across the LLMs of different scales, further illustrating its generalization ability.

\begin{table*}[t]
\centering
\small
\setlength{\fboxsep}{1.0pt}
\resizebox{\linewidth}{!}{ 
\begin{tabular}{p{1\textwidth}}
\hline
\rowcolor{gray!8}{\textbf{Case 1:} What are the virulence factors of anthrax?}\\

\hline
\textbf{Ground Truth:} \textbf{\emph{\textcolor[HTML]{EB6500}{Bacillus anthracis}}}.\\
\textbf{Choice A:} The virulence factors of anthrax are a group of proteins produced by the \textbf{\emph{\textcolor[HTML]{EB6500}{Bacillus anthracis}}} bacterium that contribute to its ability to cause disease in humans and animals.\quad \quad
\textbf{Choice B:} \textbf{\emph{\textcolor[HTML]{EB6500}{anthracis}}}. \\
\textbf{Choice C:} lethal factor, edema factor and antiphagocytic factor.\\
\textbf{Choice D:} lethal factor,antiphagocytic factor and other factors.\\\hdashline
\textbf{Raw Metric (ROUGE-L):} Choice B(0.667) > Choice A(0.143) > Choice C(0) = Choice D(0)\\
\textbf{Zero-Shot:} ...choice C and D are more complete as they specifically mention the names of the virulence factor... \textbf{\emph{\textcolor[HTML]{0084EB}{Best answer :C}}}.\\
\textbf{SFT:}... from the perspective of Coherence... \textbf{\emph{\textcolor[HTML]{0084EB}{Best answer: C}}} ... .\\
\textbf{\method{}:} from the perspective of semantic consistency, completeness, and coherence... \textbf{\emph{\textcolor{red}{Best answer: A}}} ... .\\
\hline
\rowcolor{gray!8}{\textbf{Case 2:} How much is 55\% of 40 is greater than 4/5 of 25? Options: (A) 29. (B) 776. \textbf{\emph{\textcolor[HTML]{EB6500}{(C) 2}}}. (D) 12. (E) 99.}\\
\hline
\textbf{Ground Truth:} \textbf{\emph{\textcolor[HTML]{EB6500}{(C)}}}\\
\textbf{Choice A:} According to the information of question, 55\% of 40 is 22, which is \textbf{\emph{\textcolor[HTML]{EB6500}{2}}} greater than 4/5 of 25, which equals 20. \\
\textbf{Choice B:} \textbf{\emph{\textcolor[HTML]{EB6500}{(C)}}}, 55\% of 40 is greater than 4/5 of 25 by \textbf{\emph{\textcolor[HTML]{9400FF}{1}}}.\\
\textbf{Choice C:} \textbf{\emph{\textcolor[HTML]{EB6500}{(C)}}}.555\% of 44 is smaller than 4/5 of 25 (which is 20) by \textbf{\emph{\textcolor[HTML]{EB6500}{2}}}.
\\
\textbf{Choice D:} 55\% of 40 is smaller than 4/5 of 25.\\ \hdashline
\textbf{Raw Metric (Accuracy):} Choice A(0), Choice B(1), Choice C(1), Choice D(0)\\
\textbf{Zero-Shot:} ...there is no hallucination in any of the choices.... 
\textbf{\emph{\textcolor[HTML]{0084EB}{ Choice C is the best answer}}} ... .\\
\textbf{SFT:} ... from the perspective of Completeness ..., choice C given the answer: C, ...
\textbf{\emph{\textcolor[HTML]{0084EB}{Best answer: C}}} ... .\\
\textbf{\method{}:} ... from the perspective of Semantic Consistency... \textbf{\emph{\textcolor{red}{Best answer: A }}} ... .\\
\hline
\end{tabular}}
\caption{Case Studies. If the phrase aligns with the facts, it is highlighted in \textbf{\emph{\textcolor[HTML]{EB6500}{Brown}}}, while inconsistencies are highlighted in \textbf{\emph{\textcolor[HTML]{9400FF}{Purple}}}. Correct results are marked in \textbf{\emph{\textcolor{red}{Red}}}, whereas incorrect results are highlighted in \textbf{\emph{\textcolor[HTML]{0084EB}{Blue}}}.}
\label{table1:casestudy}
\end{table*}

\subsection{Case Study}
\label{sec:5.4}
In Table~\ref{table1:casestudy}, we present two cases to illustrate the effectiveness of the \method{} method.

In the first case, the Raw Metric selects the response with the most matched phrases to the ground truth, which results in unreliable evaluation for long ground truths~\cite{EnablingLargeLanguageModelstoGenerateTextwithCitations2023GaoTianyu}. Additionally, the vanilla LLM prioritizes factual correctness, while the SFT model focuses on coherence aspects, leading to incorrect judgments. In contrast, \method{} incorporates multiple evaluation dimensions, such as ``Semantic'', ``Completeness'' and ``Coherence'', allowing for more effective and consistent evaluations. This demonstrates the effectiveness of \method{} in choosing appropriate dimensions during evaluating the responses, which helps to provide a more comprehensive assessment.

In the second case, all models select the Choice B and Choice C, which contains the ground truth answer ``(C)''. It illustrates that these models, even for these LLM-based judgment models, focus more on matching responses with the ground truth answer to conduct the judgment. However, the reasoning processes in these responses of both Choice B and Choice C contain errors. In contrast, \method{} shifts its attention from matching the ground truth answer to the ``Semantic'' dimension, showing its effectiveness.

\section{Conclusion}
This paper proposes the \method{} method, which enhances the judgment ability of LLMs in a self-improvement method. It prompts LLMs to generate different judgments based on various combinations of judgment dimensions and utilizes the consistency between judgments to enhance LLMs to generate more accurate judgments. Our experimental results show that \method{} has the ability to choose appropriate evaluation dimensions to generate reliable and comprehensive assessments.

\section*{Limitation}
Although \method{} demonstrates convincing performance in enhancing the judgment ability of LLMs, it has some limitations. First, the judgment results are often sensitive to the evaluation prompts, and \method{} relies on the consistency of these prompts to improve LLMs' judgment capability. Thus, the design of high-quality evaluation prompts that encompass comprehensive evaluation dimensions remains an underexplored area. Second, \method{} utilizes the well-performing embedding model, MiniCPM-Embedding, to calculate the similarity score between judgment results based on different evaluation aspects. However, the effectiveness of this similarity estimation method may limit the overall performance of \method{}. Further exploration of more fine-grained approaches, such as incorporating the matching signals of ground truth answers, could enhance its effectiveness.




\section*{Ethics Statement}
Our experiment involves employing human evaluators to evaluate the outputs generated by the RAG model. We ensure that we have carefully distributed the data from our experiment to the human evaluators, ensuring it is strictly used for research purposes and does not contain any content that violates ethical standards.


\bibliography{reference}

\begin{thebibliography}{79}
\providecommand{\natexlab}[1]{#1}

\bibitem[{Adlakha et~al.(2023)Adlakha, BehnamGhader, Lu, Meade, and Reddy}]{adlakha2023evaluating}
Vaibhav Adlakha, Parishad BehnamGhader, Xing~Han Lu, Nicholas Meade, and Siva Reddy. 2023.
\newblock \href {https://ArXiv.org/abs/2307.16877} {Evaluating correctness and faithfulness of instruction-following models for question answering}.
\newblock \emph{ArXiv preprint}, abs/2307.16877.

\bibitem[{Aggarwal et~al.(2021)Aggarwal, Mandowara, Agrawal, Khandelwal, Singla, and Garg}]{ecqa2021Aggarwal}
Shourya Aggarwal, Divyanshu Mandowara, Vishwajeet Agrawal, Dinesh Khandelwal, Parag Singla, and Dinesh Garg. 2021.
\newblock \href {https://aclanthology.org/2021.acl-long.238.pdf} {Explanations for commonsenseqa: New dataset and models}.
\newblock In \emph{Proceedings of ACL}, pages 3050--3065.

\bibitem[{An et~al.(2023)An, Gong, Zhong, Zhao, Li, Zhang, Kong, and Qiu}]{an2023eval}
Chenxin An, Shansan Gong, Ming Zhong, Xingjian Zhao, Mukai Li, Jun Zhang, Lingpeng Kong, and Xipeng Qiu. 2023.
\newblock \href {https://ArXiv.org/abs/2307.11088} {L-eval: Instituting standardized evaluation for long context language models}.
\newblock In \emph{Proceedings of ACL}, pages 14388--14411.

\bibitem[{Asai et~al.(2023)Asai, Wu, Wang, Sil, and Hajishirzi}]{asai2023self}
Akari Asai, Zeqiu Wu, Yizhong Wang, Avirup Sil, and Hannaneh Hajishirzi. 2023.
\newblock \href {https://ArXiv.org/abs/2310.11511} {Self-rag: Learning to retrieve, generate, and critique through self-reflection}.
\newblock In \emph{Proceedings of ICLR}.

\bibitem[{Asai et~al.(2024)Asai, Zhong, Chen, Koh, Zettlemoyer, Hajishirzi, and tau Yih}]{RALM2024Asai}
Akari Asai, Zexuan Zhong, Danqi Chen, Pang~Wei Koh, Luke Zettlemoyer, Hannaneh Hajishirzi, and Wen tau Yih. 2024.
\newblock \href {https://ArXiv.org/abs/2403.03187} {Reliable, adaptable, and attributable language models with retrieval}.
\newblock \emph{ArXiv preprint}, abs/2403.03187.

\bibitem[{Bai et~al.(2023)Bai, Bai, Chu, Cui, Dang, and Deng}]{qwen2.5-14b2023Bai}
Jinze Bai, Shuai Bai, Yunfei Chu, Zeyu Cui, Kai Dang, and Xiaodong Deng. 2023.
\newblock \href {https://ArXiv.org/abs/2309.16609} {Qwen technical report}.
\newblock \emph{ArXiv preprint}, abs/2309.16609.

\bibitem[{Bajaj et~al.(2016)Bajaj, Campos, Craswell, Deng, Gao, Liu, Majumder, McNamara, Mitra, Nguyen et~al.}]{bajaj2016ms}
Payal Bajaj, Daniel Campos, Nick Craswell, Li~Deng, Jianfeng Gao, Xiaodong Liu, Rangan Majumder, Andrew McNamara, Bhaskar Mitra, Tri Nguyen, et~al. 2016.
\newblock \href {https://ArXiv.org/abs/1611.09268} {Ms marco: A human generated machine reading comprehension dataset}.
\newblock \emph{ArXiv preprint}, abs/1611.09268.

\bibitem[{Berant et~al.(2014)Berant, Chou, Frostig, and Liang}]{webquestion2014Berant}
Jonathan Berant, Andrew Chou, Roy Frostig, and Percy Liang. 2014.
\newblock \href {https: //aclanthology.org/D13-1160} {Semantic parsing on freebase from question-answer pairs}.
\newblock In \emph{Proceedings of EMNLP}, pages 1533--1544.

\bibitem[{Cai et~al.(2019)Cai, Wang, Bi, Tu, Liu, Lam, and Shi}]{cai2019skeleton}
Deng Cai, Yan Wang, Wei Bi, Zhaopeng Tu, Xiaojiang Liu, Wai Lam, and Shuming Shi. 2019.
\newblock \href {https://aclanthology.org/N19-1124} {Skeleton-to-response: Dialogue generation guided by retrieval memory}.
\newblock In \emph{Proceedings of NAACL-HLT}, pages 1219--1228.

\bibitem[{Chan et~al.(2023)Chan, Chen, Su, Yu, Xue, Zhang, Fu, and Liu}]{chanchateval}
Chi-Min Chan, Weize Chen, Yusheng Su, Jianxuan Yu, Wei Xue, Shanghang Zhang, Jie Fu, and Zhiyuan Liu. 2023.
\newblock \href {https://ArXiv.org/abs/2308.07201} {Chateval: Towards better llm-based evaluators through multi-agent debate}.
\newblock In \emph{Proceedings of ICLR}.

\bibitem[{Chen et~al.(2024)Chen, Chen, Liu, Jiang, and Wang}]{chen2024humans}
Guiming~Hardy Chen, Shunian Chen, Ziche Liu, Feng Jiang, and Benyou Wang. 2024.
\newblock \href {https://aclanthology.org/2024.emnlp-main.474/} {Humans or llms as the judge? a study on judgement biases}.
\newblock In \emph{Proceedings of EMNLP}, pages 8301--8327.

\bibitem[{Chen et~al.(2023{\natexlab{a}})Chen, Wang, Ji, Gao, Jiang, Chen, Zhang, Song, Xie, Kong et~al.}]{chen2023huatuogpt}
Junying Chen, Xidong Wang, Ke~Ji, Anningzhe Gao, Feng Jiang, Shunian Chen, Hongbo Zhang, Dingjie Song, Wenya Xie, Chuyi Kong, et~al. 2023{\natexlab{a}}.
\newblock \href {https://arxiv.org/abs/2311.09774} {Huatuogpt-ii, one-stage training for medical adaption of llms}.
\newblock \emph{ArXiv preprint}, abs/2311.09774.

\bibitem[{Chen et~al.(2023{\natexlab{b}})Chen, Lin, Schaerli, and Zhou}]{chen2023teaching}
Xinyun Chen, Maxwell Lin, Nathanael Schaerli, and Denny Zhou. 2023{\natexlab{b}}.
\newblock \href {https://ArXiv.org/abs/2304.05128} {Teaching large language models to self-debug}.
\newblock In \emph{Proceedings of ICLR}.

\bibitem[{Chen et~al.(2021)Chen, Chen, Smiley, Shah, Borova, Langdon, Moussa, Beane, Huang, Routledge et~al.}]{chen2021finqa}
Zhiyu Chen, Wenhu Chen, Charese Smiley, Sameena Shah, Iana Borova, Dylan Langdon, Reema Moussa, Matt Beane, Ting-Hao Huang, Bryan~R Routledge, et~al. 2021.
\newblock \href {https://ArXiv.org/abs/2109.00122} {Finqa: A dataset of numerical reasoning over financial data}.
\newblock In \emph{Proceedings of EMNLP}, pages 3697--3711.

\bibitem[{Chiang and Lee(2023{\natexlab{a}})}]{chiang2023can}
Cheng-Han Chiang and Hung-Yi Lee. 2023{\natexlab{a}}.
\newblock \href {https://ArXiv.org/abs/2305.01937} {Can large language models be an alternative to human evaluations?}
\newblock In \emph{Proceedings of ACL}, pages 15607--15631.

\bibitem[{Chiang and Lee(2023{\natexlab{b}})}]{chiang2023closer}
Cheng-Han Chiang and Hung-yi Lee. 2023{\natexlab{b}}.
\newblock \href {https://ArXiv.org/abs/2310.05657} {A closer look into automatic evaluation using large language models}.
\newblock In \emph{Proceedings of EMNLP Findings}, pages 8928--8942.

\bibitem[{Chung et~al.(2022)Chung, Hou, Longpre, Zoph, Tay, Fedus, and Li}]{Chung2022flan_t5}
Hyung~Won Chung, Le~Hou, Shayne Longpre, Barret Zoph, Yi~Tay, William Fedus, and Yunxuan Li. 2022.
\newblock \href {https://ArXiv.org/abs/2210.11416} {Scaling instruction-finetuned language models}.
\newblock \emph{ArXiv preprint}, abs/2210.11416.

\bibitem[{Dinan et~al.(2019)Dinan, Roller, Shuster, Fan, Auli, and Weston}]{wow2019Dinan}
Emily Dinan, Stephen Roller, Kurt Shuster, Angela Fan, Michael Auli, and Jason Weston. 2019.
\newblock \href {https://openreview.net/forum?id=r1l73iRqKm} {Wizard of wikipedia: Knowledge-powered conversational agents}.
\newblock In \emph{Proceedings of ICLR}.

\bibitem[{Du et~al.(2022)Du, Qian, Liu, Ding, Qiu, Yang, and Tang}]{du2022glm}
Zhengxiao Du, Yujie Qian, Xiao Liu, Ming Ding, Jiezhong Qiu, Zhilin Yang, and Jie Tang. 2022.
\newblock \href {https://ArXiv.org/abs/2103.10360} {Glm: General language model pretraining with autoregressive blank infilling}.
\newblock In \emph{Proceedings of ACL}, pages 320--335.

\bibitem[{Elazar et~al.(2021)Elazar, Kassner, Ravfogel, Ravichander, Hovy, Sch{\"u}tze, and Goldberg}]{elazar2021measuring}
Yanai Elazar, Nora Kassner, Shauli Ravfogel, Abhilasha Ravichander, Eduard Hovy, Hinrich Sch{\"u}tze, and Yoav Goldberg. 2021.
\newblock \href {https://ArXiv.org/abs/2102.01017} {Measuring and improving consistency in pretrained language models}.
\newblock \emph{Transactions of the Association for Computational Linguistics}, pages 1012--1031.

\bibitem[{Fan et~al.(2019)Fan, Jernite, Perez, Grangier, Weston, and Auli}]{fan-etal-2019-eli5}
Angela Fan, Yacine Jernite, Ethan Perez, David Grangier, Jason Weston, and Michael Auli. 2019.
\newblock \href {https://aclanthology.org/P19-1346/} {{ELI}5: Long form question answering}.
\newblock In \emph{Proceedings of ACL}, pages 3558--3567.

\bibitem[{Friel et~al.(2024)Friel, Belyi, and Sanyal}]{friel2024ragbench}
Robert Friel, Masha Belyi, and Atindriyo Sanyal. 2024.
\newblock \href {https://ArXiv.org/abs/2407.11005} {Ragbench: Explainable benchmark for retrieval-augmented generation systems}.
\newblock \emph{ArXiv preprint}, abs/2407.11005.

\bibitem[{Gao et~al.(2023{\natexlab{a}})Gao, Yen, Yu, and Chen}]{EnablingLargeLanguageModelstoGenerateTextwithCitations2023GaoTianyu}
Tianyu Gao, Howard Yen, Jiatong Yu, and Danqi Chen. 2023{\natexlab{a}}.
\newblock \href {https://ArXiv.org/pdf/2305.14627} {Enabling large language models to generate text with citations}.
\newblock In \emph{Proceedings of EMNLP}, pages 6465--6488.

\bibitem[{Gao et~al.(2023{\natexlab{b}})Gao, Xiong, Gao, Jia, Pan, Bi, Dai, Sun, Wang, and Wang}]{Retrieval2023Gao}
Yunfan Gao, Yun Xiong, Xinyu Gao, Kangxiang Jia, Jinliu Pan, Yuxi Bi, Yi~Dai, Jiawei Sun, Meng Wang, and Haofen Wang. 2023{\natexlab{b}}.
\newblock \href {https://ArXiv.org/abs/2312.10997} {Retrieval-augmented generation for large language models: A survey}.
\newblock \emph{ArXiv preprint}, abs/2312.10997.

\bibitem[{Geva et~al.(2021)Geva, Khashabi, Segal, Khot, Roth, and Berant}]{strategyqa2021Geva}
Mor Geva, Daniel Khashabi, Elad Segal, Tushar Khot, Dan Roth, and Jonathan Berant. 2021.
\newblock \href {https://ArXiv.org/abs/2101.02235} {Did aristotle use a laptop? a question answering benchmark with implicit reasoning strategies}.
\newblock In \emph{Transactions of the Association for Computational Linguistics}, pages 346--361.

\bibitem[{Gu et~al.(2024)Gu, Jiang, Shi, and Tan}]{SurveyonLLM-as-a-Judge2024Gu}
Jiawei Gu, Xuhui Jiang, Zhichao Shi, and Hexiang Tan. 2024.
\newblock \href {https://ArXiv.org/abs/2411.15594} {A survey on llm-as-a-judge}.
\newblock \emph{ArXiv preprint}, abs/2411.15594.

\bibitem[{Guu et~al.(2020)Guu, Lee, Tung, Pasupat, and Chang}]{REALM2020Guu}
Kelvin Guu, Kenton Lee, Zora Tung, Panupong Pasupat, and Ming-Wei Chang. 2020.
\newblock \href {https://ArXiv.org/abs/2002.08909} {Realm: Retrieval-augmented language model pre-training}.
\newblock In \emph{Proceedings of ICML}, pages 3929--3938.

\bibitem[{He et~al.(2021)He, Neubig, and Berg-Kirkpatrick}]{he2021efficient}
Junxian He, Graham Neubig, and Taylor Berg-Kirkpatrick. 2021.
\newblock \href {https://ArXiv.org/pdf/2109.04212} {Efficient nearest neighbor language models}.
\newblock In \emph{Proceedings of EMNLP}, pages 5703--5714.

\bibitem[{Hu et~al.(2022)Hu, Wallis, Allen-Zhu, Li, Wang, Wang, Chen et~al.}]{hulora}
Edward~J Hu, Phillip Wallis, Zeyuan Allen-Zhu, Yuanzhi Li, Shean Wang, Lu~Wang, Weizhu Chen, et~al. 2022.
\newblock \href {https://ArXiv.org/pdf/2106.09685} {Lora: Low-rank adaptation of large language models}.
\newblock In \emph{Proceedings of ICLR}.

\bibitem[{Hu et~al.(2024)Hu, Tu, Han, He, Cui, and Long}]{minicpm-2b2024Hu}
Shengding Hu, Yuge Tu, Xu~Han, Chaoqun He, Ganqu Cui, and Xiang Long. 2024.
\newblock \href {https://ArXiv.org/abs/2404.06395} {Minicpm: Unveiling the potential of small language models with scalable training strategies}.
\newblock \emph{ArXiv preprint}, abs/2404.06395.

\bibitem[{Huang et~al.(2023)Huang, Yu, Ma, Zhong, Feng, Wang, Chen, Peng, Feng, and Qin}]{huang2023survey}
Lei Huang, Weijiang Yu, Weitao Ma, Weihong Zhong, Zhangyin Feng, Haotian Wang, Qianglong Chen, Weihua Peng, Xiaocheng Feng, and Bing Qin. 2023.
\newblock \href {https://ArXiv.org/abs/2311.05232} {A survey on hallucination in large language models: Principles, taxonomy, challenges, and open questions}.
\newblock \emph{ACM Transactions on Information Systems}.

\bibitem[{Izacard et~al.(2022)Izacard, Lewis, Lomeli, Hosseini, Petroni, Schick, Dwivedi-Yu, Joulin, Riedel, and Grave}]{izacard2022few}
Gautier Izacard, Patrick Lewis, Maria Lomeli, Lucas Hosseini, Fabio Petroni, Timo Schick, Jane Dwivedi-Yu, Armand Joulin, Sebastian Riedel, and Edouard Grave. 2022.
\newblock \href {https://ArXiv.org/abs/2208.03299} {Atlas: few-shot learning with retrieval augmented language models}.
\newblock \emph{ArXiv preprint}, abs/2208.03299.

\bibitem[{Jacovi et~al.(2025)Jacovi, Wang, Alberti, Tao, Lipovetz, Olszewska, Haas, Liu, Keating, Bloniarz et~al.}]{jacovi2025facts}
Alon Jacovi, Andrew Wang, Chris Alberti, Connie Tao, Jon Lipovetz, Kate Olszewska, Lukas Haas, Michelle Liu, Nate Keating, Adam Bloniarz, et~al. 2025.
\newblock \href {https://ArXiv.org/abs/2501.03200} {The facts grounding leaderboard: Benchmarking llms' ability to ground responses to long-form input}.
\newblock \emph{ArXiv preprint}, abs/2501.03200.

\bibitem[{Ji et~al.(2023)Ji, Lee, Frieske, Yu, Su, Xu, Ishii, Bang, Madotto, and Fung}]{ji2023survey}
Ziwei Ji, Nayeon Lee, Rita Frieske, Tiezheng Yu, Dan Su, Yan Xu, Etsuko Ishii, Ye~Jin Bang, Andrea Madotto, and Pascale Fung. 2023.
\newblock \href {https://ArXiv.org/pdf/2202.03629} {Survey of hallucination in natural language generation}.
\newblock \emph{ACM Computing Surveys}.

\bibitem[{Jin et~al.(2019)Jin, Dhingra, Liu, Cohen, and Lu}]{jin2019pubmedqa}
Qiao Jin, Bhuwan Dhingra, Zhengping Liu, William Cohen, and Xinghua Lu. 2019.
\newblock \href {https://aclanthology.org/D19-1259/} {Pubmedqa: A dataset for biomedical research question answering}.
\newblock In \emph{Proceedings of EMNLP-IJCNLP}, pages 2567--2577.

\bibitem[{Jin et~al.(2024)Jin, Yuan, Men, Cao, Chen, Liu, and Zhao}]{jin2024rag}
Zhuoran Jin, Hongbang Yuan, Tianyi Men, Pengfei Cao, Yubo Chen, Kang Liu, and Jun Zhao. 2024.
\newblock \href {https://ArXiv.org/abs/2412.13746} {Rag-rewardbench: Benchmarking reward models in retrieval augmented generation for preference alignment}.
\newblock \emph{ArXiv preprint}, abs/2412.13746.

\bibitem[{Joshi et~al.(2017)Joshi, Choi, Weld, and Zettlemoyer}]{triviaqa2017Joshi}
Mandar Joshi, Eunsol Choi, Daniel Weld, and Luke Zettlemoyer. 2017.
\newblock \href {https://aclanthology.org/P17-1147.} {Triviaqa: A large scale distantly supervised challenge dataset for reading comprehension}.
\newblock In \emph{Proceedings of ACL}, pages 1601--1611.

\bibitem[{Kim et~al.(2023)Kim, Shin, Cho, Jang, Longpre, Lee, Yun, Shin, Kim, Thorne et~al.}]{kim2023prometheus}
Seungone Kim, Jamin Shin, Yejin Cho, Joel Jang, Shayne Longpre, Hwaran Lee, Sangdoo Yun, Seongjin Shin, Sungdong Kim, James Thorne, et~al. 2023.
\newblock \href {https://ArXiv.org/abs/2310.08491} {Prometheus: Inducing fine-grained evaluation capability in language models}.
\newblock In \emph{Proceedings of ICLR}.

\bibitem[{Ko{\v{c}}isk{\`y} et~al.(2018)Ko{\v{c}}isk{\`y}, Schwarz, Blunsom, Dyer, Hermann, Melis, and Grefenstette}]{kovcisky2018narrativeqa}
Tom{\'a}{\v{s}} Ko{\v{c}}isk{\`y}, Jonathan Schwarz, Phil Blunsom, Chris Dyer, Karl~Moritz Hermann, G{\'a}bor Melis, and Edward Grefenstette. 2018.
\newblock \href {https://aclanthology.org/Q18-1023/} {The narrativeqa reading comprehension challenge}.
\newblock \emph{Transactions of the Association for Computational Linguistics}, pages 317--328.

\bibitem[{Kwiatkowski et~al.(2019)Kwiatkowski, Palomaki, Redfield, Collins, Parikh, and Alberti}]{nq2019Kwiatkowski}
Tom Kwiatkowski, Jennimaria Palomaki, Olivia Redfield, Michael Collins, Ankur Parikh, and Chris Alberti. 2019.
\newblock \href {https://research.google/pubs/natural-questions-a-benchmark-for-question-answering-research/} {Natural questions: a benchmark for question answering research}.
\newblock In \emph{Proceedings of ACL}, pages 452--466.

\bibitem[{Lewis et~al.(2020)Lewis, Perez, Piktus, Petroni, Karpukhin, Goyal, Küttler, Lewis, tau Yih, Rocktäschel, Riedel, and Kiela}]{Retrieval2020Lewis}
Patrick S.~H. Lewis, Ethan Perez, Aleksandra Piktus, Fabio Petroni, Vladimir Karpukhin, Naman Goyal, Heinrich Küttler, Mike Lewis, Wen tau Yih, Tim Rocktäschel, Sebastian Riedel, and Douwe Kiela. 2020.
\newblock \href {https://ArXiv.org/abs/2005.11401} {Retrieval-augmented generation for knowledge-intensive nlp tasks}.
\newblock In \emph{Proceedings of NeurIPS}, pages 9459--9474.

\bibitem[{Li et~al.(2024{\natexlab{a}})Li, Dong, Chen, Su, Zhou, Ai, Ye, and Liu}]{li2024llms}
Haitao Li, Qian Dong, Junjie Chen, Huixue Su, Yujia Zhou, Qingyao Ai, Ziyi Ye, and Yiqun Liu. 2024{\natexlab{a}}.
\newblock \href {https://ArXiv.org/abs/2412.05579} {Llms-as-judges: A comprehensive survey on llm-based evaluation methods}.
\newblock \emph{ArXiv preprint}, abs/2412.05579.

\bibitem[{Li et~al.(2023)Li, Yang, Cheng, Liu, Yu, Yang, and Lam}]{self-improve2023Li}
Siheng Li, Cheng Yang, Zesen Cheng, Lemao Liu, Mo~Yu, Yujiu Yang, and Wai Lam. 2023.
\newblock \href {https://ArXiv.org/abs/2411.08147} {Large language models can self-improve in long-context reasoning}.
\newblock In \emph{Proceedings of EMNLP}, pages 1051--1068.

\bibitem[{Li et~al.(2024{\natexlab{b}})Li, Mei, Liu, Yan, Wang, Yu, Zeng, Chen, Yu, and Liu}]{rag-ddr2024Li}
Xinze Li, Sen Mei, Zhenghao Liu, Yukun Yan, Shuo Wang, Shi Yu, Zheni Zeng, Hao Chen, Ge~Yu, and Zhiyuan Liu. 2024{\natexlab{b}}.
\newblock \href {https://ArXiv.org/abs/2410.13509} {Rag-ddr: Optimizing retrieval-augmented generation using differentiable data rewards}.
\newblock \emph{ArXiv preprint}, abs/2410.13509.

\bibitem[{Li et~al.(2025)Li, Wang, Liu, Yu, Wang, Yan, Fu, Gu, and Yu}]{li2025building}
Xinze Li, Hanbin Wang, Zhenghao Liu, Shi Yu, Shuo Wang, Yukun Yan, Yukai Fu, Yu~Gu, and Ge~Yu. 2025.
\newblock \href {https://ArXiv.org/abs/2410.16229} {Building a coding assistant via the retrieval-augmented language model}.
\newblock \emph{ACM Transactions on Information Systems}, pages 1--25.

\bibitem[{Lin et~al.(2023)Lin, Chen, Chen, Shi, Lomeli, James, Rodriguez, Kahn, Szilvasy, and Lewis}]{Radit2023Lin}
Xi~Victoria Lin, Xilun Chen, Mingda Chen, Weijia Shi, Maria Lomeli, Rich James, Pedro Rodriguez, Jacob Kahn, Gergely Szilvasy, and Mike Lewis. 2023.
\newblock \href {https://ArXiv.org/abs/2310.01352} {Ra-dit: Retrieval-augmented dual instruction tuning}.
\newblock In \emph{Proceedings of ICLR}.

\bibitem[{Ling et~al.(2017)Ling, Yogatama, Dyer, and Blunsom}]{aquarat2017Ling}
Wang Ling, Dani Yogatama, Chris Dyer, and Phil Blunsom. 2017.
\newblock \href {https://aclanthology.org/P17-1015/} {Program induction by rationale generation : Learning to solve and explain algebraic word problems}.
\newblock In \emph{Proceedings of ACL}, pages 158--167.

\bibitem[{Liu et~al.(2024)Liu, Zhou, Guo, Shareghi, Vuli{\'c}, Korhonen, and Collier}]{liu2024aligning}
Yinhong Liu, Han Zhou, Zhijiang Guo, Ehsan Shareghi, Ivan Vuli{\'c}, Anna Korhonen, and Nigel Collier. 2024.
\newblock \href {https://ArXiv.org/abs/2403.16950} {Aligning with human judgement: The role of pairwise preference in large language model evaluators}.
\newblock In \emph{Proceedings of COLM}.

\bibitem[{Lu et~al.(2022)Lu, Mishra, Xia, Qiu, Chang, Zhu, Tafjord, Clark, and Kalyan}]{lu2022learn}
Pan Lu, Swaroop Mishra, Tony Xia, Liang Qiu, Kai-Wei Chang, Song-Chun Zhu, Oyvind Tafjord, Peter Clark, and Ashwin Kalyan. 2022.
\newblock \href {https://ArXiv.org/abs/2209.09513} {Learn to explain: Multimodal reasoning via thought chains for science question answering}.
\newblock In \emph{Proceedings of NeurIPS}, pages 2507--2521.

\bibitem[{Maia et~al.(2018)Maia, Handschuh, Freitas, Davis, McDermott, Zarrouk, and Balahur}]{maia201818}
Macedo Maia, Siegfried Handschuh, Andr{\'e} Freitas, Brian Davis, Ross McDermott, Manel Zarrouk, and Alexandra Balahur. 2018.
\newblock \href {https://dl.acm.org/doi/10.1145/3184558.3192301} {Www'18 open challenge: financial opinion mining and question answering}.
\newblock In \emph{proceedings of the the web conference 2018}, pages 1941--1942.

\bibitem[{Mallen et~al.(2023)Mallen, Asai, Zhong, Das, Khashabi, and Hajishirzi}]{mallen2023not}
Alex Mallen, Akari Asai, Victor Zhong, Rajarshi Das, Daniel Khashabi, and Hannaneh Hajishirzi. 2023.
\newblock \href {https://ArXiv.org/abs/2212.10511} {When not to trust language models: Investigating effectiveness of parametric and non-parametric memories}.
\newblock In \emph{Proceedings of ACL}, pages 9802--9822.

\bibitem[{Niu et~al.(2024)Niu, Joty, Liu, Xiong, Zhou, and Yavuz}]{niu2024judgerank}
Tong Niu, Shafiq Joty, Ye~Liu, Caiming Xiong, Yingbo Zhou, and Semih Yavuz. 2024.
\newblock \href {https://ArXiv.org/abs/2411.00142} {Judgerank: Leveraging large language models for reasoning-intensive reranking}.
\newblock \emph{ArXiv preprint}, abs/2411.00142.

\bibitem[{OpenAI(2023)}]{openai2023gpt}
R~OpenAI. 2023.
\newblock \href {https://doi.org/10.48550/ArXiv.2303.08774} {Gpt-4 technical report}.
\newblock \emph{ArXiv preprint}, abs/2303.08774.

\bibitem[{Rafailov et~al.(2023)Rafailov, Sharma, Mitchell, Ermon, Manning, and Finn}]{DPO2023Rafailov}
Rafael Rafailov, Archit Sharma, Eric Mitchell, Stefano Ermon, Christopher~D. Manning, and Chelsea Finn. 2023.
\newblock \href {https://ArXiv.org/abs/2305.18290} {Direct preference optimization: Your language model is secretly a reward model}.
\newblock In \emph{Proceedings of NeurIPS}, pages 53728--53741.

\bibitem[{Ram et~al.(2023)Ram, Levine, Dalmedigos, Muhlgay, Shashua, Leyton-Brown, , and Shoham}]{Ram2023Incontextlearning}
Ori Ram, Yoav Levine, Itay Dalmedigos, Dor Muhlgay, Amnon Shashua, Kevin Leyton-Brown, , and Yoav Shoham. 2023.
\newblock \href {https://ArXiv.org/abs/2302.00083} {In-context retrieval-augmented language models}.
\newblock \emph{Transactions of the Association for Computational Linguistics}.

\bibitem[{Saad-Falcon et~al.(2024)Saad-Falcon, Khattab, Potts, and Zaharia}]{saad2024ares}
Jon Saad-Falcon, Omar Khattab, Christopher Potts, and Matei Zaharia. 2024.
\newblock \href {https://ArXiv.org/abs/2311.09476} {Ares: An automated evaluation framework for retrieval-augmented generation systems}.
\newblock In \emph{Proceedings of NAACL}, pages 338--354.

\bibitem[{Shi et~al.(2023)Shi, Min, Yasunaga, Seo, James, Lewis, Zettlemoyer, and Yih}]{shi2023replug}
Weijia Shi, Sewon Min, Michihiro Yasunaga, Minjoon Seo, Rich James, Mike Lewis, Luke Zettlemoyer, and Wen-tau Yih. 2023.
\newblock \href {https://ArXiv.org/abs/2301.12652} {Replug: Retrieval-augmented black-box language models}.
\newblock In \emph{Proceedings of NAACL}, pages 8371--8384.

\bibitem[{Shuster et~al.(2021)Shuster, Poff, Chen, Kiela, and Weston}]{shuster2021retrieval}
Kurt Shuster, Spencer Poff, Moya Chen, Douwe Kiela, and Jason Weston. 2021.
\newblock \href {https://aclanthology.org/2021.findings-emnlp.320/} {Retrieval augmentation reduces hallucination in conversation}.
\newblock In \emph{Proceedings of EMNLP Findings}, pages 3784--3803.

\bibitem[{Sottana et~al.(2023)Sottana, Liang, Zou, and Yuan}]{sottana2023evaluation}
Andrea Sottana, Bin Liang, Kai Zou, and Zheng Yuan. 2023.
\newblock \href {https://ArXiv.org/abs/2310.13800} {Evaluation metrics in the era of gpt-4: Reliably evaluating large language models on sequence to sequence tasks}.
\newblock In \emph{Proceedings of EMNLP}, pages 8776--8788.

\bibitem[{Stelmakh et~al.(2022)Stelmakh, Luan, Dhingra, and Chang}]{ASQA2022Stelmakh}
Ivan Stelmakh, Yi~Luan, Bhuwan Dhingra, and Ming-Wei Chang. 2022.
\newblock \href {https://aclanthology.org/2022.emnlp-main.566/} {Asqa: Factoid questions meet long-form answers}.
\newblock In \emph{Proceedings of EMNLP}, pages 8273--8288.

\bibitem[{Suri et~al.(2021)Suri, Zhang, Huo, Liu, and Guan}]{suri2021mediaqa}
Huqun Suri, Qi~Zhang, Wenhua Huo, Yan Liu, and Chunsheng Guan. 2021.
\newblock \href {https://ArXiv.org/abs/2108.08074} {Mediaqa: A question answering dataset on medical dialogues}.
\newblock \emph{ArXiv preprint}, abs/2108.08074.

\bibitem[{Touvron et~al.(2023)Touvron, Martin, Stone, Albert, Almahairi, Babaei, Bashlykov, Batra, Bhargava, Bhosale et~al.}]{touvron2023llama}
Hugo Touvron, Louis Martin, Kevin Stone, Peter Albert, Amjad Almahairi, Yasmine Babaei, Nikolay Bashlykov, Soumya Batra, Prajjwal Bhargava, Shruti Bhosale, et~al. 2023.
\newblock \href {https://ArXiv.org/abs/2307.09288} {Llama 2: Open foundation and fine-tuned chat models}.
\newblock \emph{ArXiv preprint}, abs/2307.09288.

\bibitem[{Trivedi et~al.(2022)Trivedi, Balasubramanian, Khot, and Sabharwal}]{trivedi2022musique}
Harsh Trivedi, Niranjan Balasubramanian, Tushar Khot, and Ashish Sabharwal. 2022.
\newblock \href {https://ArXiv.org/abs/2108.00573} {Musique: Multihop questions via single-hop question composition}.
\newblock \emph{Transactions of the Association for Computational Linguistics}, pages 539--554.

\bibitem[{Trivedi et~al.(2023)Trivedi, Balasubramanian, Khot, and Sabharwal}]{trivedi2023interleaving}
Harsh Trivedi, Niranjan Balasubramanian, Tushar Khot, and Ashish Sabharwal. 2023.
\newblock \href {https://aclanthology.org/2023.acl-long.557.pdf} {Interleaving retrieval with chain-of-thought reasoning for knowledge-intensive multi-step questions}.
\newblock In \emph{Proceedings of ACL}, pages 10014--10037.

\bibitem[{Wang et~al.(2023{\natexlab{a}})Wang, Zhou, Chang, Liu, Zhang, Du, Xiao, and Zhu}]{wang2023learning}
Chenglong Wang, Hang Zhou, Kaiyan Chang, Tongran Liu, Chunliang Zhang, Quan Du, Tong Xiao, and Jingbo Zhu. 2023{\natexlab{a}}.
\newblock \href {https://ArXiv.org/abs/2308.04386} {Learning evaluation models from large language models for sequence generation}.
\newblock \emph{ArXiv preprint}, abs/2308.04386.

\bibitem[{Wang et~al.(2023{\natexlab{b}})Wang, Yu, Yao, Zeng, Yang, Wang, Chen, Jiang, Xie, Wang et~al.}]{wangpandalm}
Yidong Wang, Zhuohao Yu, Wenjin Yao, Zhengran Zeng, Linyi Yang, Cunxiang Wang, Hao Chen, Chaoya Jiang, Rui Xie, Jindong Wang, et~al. 2023{\natexlab{b}}.
\newblock \href {https://ArXiv.org/abs/2306.05087} {Pandalm: An automatic evaluation benchmark for llm instruction tuning optimization}.
\newblock In \emph{Proceedings of ICLR}.

\bibitem[{Wei et~al.(2022)Wei, Wang, Schuurmans, Bosma, Ichter, Xia, Chi, Le, and Zhou}]{Chain-of-Thought2022Wei}
Jason Wei, Xuezhi Wang, Dale Schuurmans, Maarten Bosma, Brian Ichter, Fei Xia, Ed~Chi, Quoc Le, and Denny Zhou. 2022.
\newblock \href {https://ArXiv.org/abs/2201.11903} {Chain-of-thought prompting elicits reasoning in large language models}.
\newblock In \emph{Proceedings of NeurIPS}, pages 24824--24837.

\bibitem[{Wei et~al.(2024)Wei, Yan, Lu, Zhu, Wang, and Zhang}]{wei2024cnnsum}
Lingxiao Wei, He~Yan, Xiangju Lu, Junmin Zhu, Jun Wang, and Wei Zhang. 2024.
\newblock \href {https://ArXiv.org/abs/2412.02819} {Cnnsum: Exploring long-conext summarization with large language models in chinese novels}.
\newblock \emph{ArXiv preprint}, abs/2412.02819.

\bibitem[{Xiao et~al.(2023)Xiao, Liu, Zhang, and Muennighoff}]{bge_embedding}
Shitao Xiao, Zheng Liu, Peitian Zhang, and Niklas Muennighoff. 2023.
\newblock \href {https://arxiv.org/abs/2309.07597} {C-pack: Packaged resources to advance general chinese embedding}.
\newblock \emph{Preprint}, ArXiv:2309.07597.

\bibitem[{Xu et~al.(2024{\natexlab{a}})Xu, Pang, Yu, Meng, Shen, Cheng, and Zhou}]{xu2024unsupervised}
Shicheng Xu, Liang Pang, Mo~Yu, Fandong Meng, Huawei Shen, Xueqi Cheng, and Jie Zhou. 2024{\natexlab{a}}.
\newblock \href {https://ArXiv.org/abs/2402.18150} {Unsupervised information refinement training of large language models for retrieval-augmented generation}.
\newblock \emph{ArXiv preprint}, abs/2402.18150.

\bibitem[{Xu et~al.(2024{\natexlab{b}})Xu, Jain, and Kankanhalli}]{xu2024hallucination}
Ziwei Xu, Sanjay Jain, and Mohan Kankanhalli. 2024{\natexlab{b}}.
\newblock \href {https://ArXiv.org/abs/2401.11817} {Hallucination is inevitable: An innate limitation of large language models}.
\newblock \emph{ArXiv preprint}, abs/2401.11817.

\bibitem[{Yan et~al.(2024)Yan, Qin, Zhuang, Jagerman, Wang, Bendersky, and Oosterhuis}]{yan2024consolidating}
Le~Yan, Zhen Qin, Honglei Zhuang, Rolf Jagerman, Xuanhui Wang, Michael Bendersky, and Harrie Oosterhuis. 2024.
\newblock \href {https://ArXiv.org/abs/2404.11791} {Consolidating ranking and relevance predictions of large language models through post-processing}.
\newblock \emph{ArXiv preprint}, abs/2404.11791.

\bibitem[{Yang et~al.(2015)Yang, tau Yih, and Meek}]{wikiqa2015Bajaj}
Yi~Yang, Wen tau Yih, and Christopher Meek. 2015.
\newblock \href {https://aclanthology.org/D15-1237/} {Wikiqa: A challenge dataset for open-domain question answering}.
\newblock In \emph{Proceedings of EMNLP}, pages 2013--2018.

\bibitem[{Yang et~al.(2018)Yang, Qi, Zhang, Bengio, Cohen, Salakhutdinov, and Manning}]{hotpotqa2018Yang}
Zhilin Yang, Peng Qi, Saizheng Zhang, Yoshua Bengio, William Cohen, Ruslan Salakhutdinov, and Christopher~D. Manning. 2018.
\newblock \href {https://aclanthology. org/D18-1259.} {Hotpotqa: A dataset for diverse, explainable multi-hop question answering}.
\newblock In \emph{Transactions of the Association for Computational Linguistics}, pages 2369--2380.

\bibitem[{Zhang et~al.(2025)Zhang, Song, Zhu, Wu, Zhang, and Niu}]{zhang2025rag}
Hanning Zhang, Juntong Song, Juno Zhu, Yuanhao Wu, Tong Zhang, and Cheng Niu. 2025.
\newblock \href {https://ArXiv.org/abs/2501.13264} {Rag-reward: Optimizing rag with reward modeling and rlhf}.
\newblock \emph{ArXiv preprint}, abs/2501.13264.

\bibitem[{Zhang et~al.(2024)Zhang, Zhang, Yuan, Liu, Shi, Gui, Zhang, and Huang}]{llmeval2024zhang}
Yue Zhang, Ming Zhang, Haipeng Yuan, Shichun Liu, Yongyao Shi, Tao Gui, Qi~Zhang, and Xuanjing Huang. 2024.
\newblock \href {https://ArXiv.org/abs/2312.07398} {Llmeval: A preliminary study on how to evaluate large language models}.
\newblock In \emph{Proceedings of AAAI}, pages 19615--19622.

\bibitem[{Zheng et~al.(2023)Zheng, Chiang, Sheng, Zhuang, Wu, Zhuang, Lin, Li, Li, Xing et~al.}]{zheng2023judging}
Lianmin Zheng, Wei-Lin Chiang, Ying Sheng, Siyuan Zhuang, Zhanghao Wu, Yonghao Zhuang, Zi~Lin, Zhuohan Li, Dacheng Li, Eric~P Xing, et~al. 2023.
\newblock \href {https://ArXiv.org/abs/2306.05685} {Judging llm-as-a-judge with mt-bench and chatbot arena}.
\newblock In \emph{Proceedings of NeurIPS}, pages 46595--46623.

\bibitem[{Zhou et~al.(2023)Zhou, Wan, Vuli{\'c}, and Korhonen}]{zhou2023survival}
Han Zhou, Xingchen Wan, Ivan Vuli{\'c}, and Anna Korhonen. 2023.
\newblock \href {https://ArXiv.org/abs/2310.12774} {Survival of the most influential prompts: Efficient black-box prompt search via clustering and pruning}.
\newblock In \emph{Proceedings of EMNLP Findings}, pages 13064--13077.

\bibitem[{Zhu et~al.(2024)Zhu, Luo, Xu, Wang, Yu, Wang, Yan, Liu, Han, Liu, and Sun}]{Rageval2024Zhu}
Kunlun Zhu, Yifan Luo, Dingling Xu, Ruobing Wang, Shi Yu, Shuo Wang, Yukun Yan, Zhenghao Liu, Xu~Han, Zhiyuan Liu, and Maosong Sun. 2024.
\newblock \href {https://ArXiv.org/abs/2408.01262} {Rageval: Scenario specific rag evaluation dataset generation framework}.
\newblock \emph{ArXiv preprint}, abs/2408.01262.

\end{thebibliography}

\clearpage
\appendix
\section{Appendix}
\label{sec:appendix}

\begin{table}[t]
\centering
\small
\begin{tabular}{l|r|r|r}
\hline
\textbf{Dataset} & \textbf{Total} & \textbf{Train} & \textbf{Dev}\\
\hline
\rowcolor{gray!8}\multicolumn{4}{l}{\textit{\method{} Training Data}}\\
\hline
FinQA~\shortcite{chen2021finqa} &4,000 &3,800 &200\\
FiQA~\shortcite{maia201818} & 8,000 &7,600 &400\\
MeDiaQA~\shortcite{suri2021mediaqa} & 2,000 &1,900 &100\\
PubMedQA~\shortcite{jin2019pubmedqa}& 2,000 &1,900 &100 \\
ScienceQA~\shortcite{lu2022learn} & 7,095 &6,740 &355\\
NQ~\shortcite{nq2019Kwiatkowski}& 4,000 & 3,800 & 200 \\
ELI5~\shortcite{fan-etal-2019-eli5} & 4,000 &3,800 &200\\
NarrativeQA~\shortcite{kovcisky2018narrativeqa} & 2,000 &1,900 &100 \\
PopQA~\shortcite{mallen2023not} & 14,267 &13,554 & 713\\
CNNSum~\shortcite{wei2024cnnsum} & 8,000 &7,600 &400\\
MuSiQue~\shortcite{trivedi2022musique} & 22,355 &21,237 &1,118\\
\hline
\rowcolor{gray!8}\multicolumn{4}{l}{\textit{RAG Training Data}}\\
\hline
ECQA~\shortcite{ecqa2021Aggarwal}&4,200 &4,000 &200 \\
MARCOQA~\shortcite{bajaj2016ms}&4,200 &4,000 &200 \\
Web Questions~\shortcite{webquestion2014Berant}&3,778 &3,578 &200 \\
WiKiQA~\shortcite{wikiqa2015Bajaj}&1,040 &840 &200 \\

Yahoo!QA&4,200 &4,000 &200 \\
StrategyQA~\shortcite{strategyqa2021Geva}&2,060 &1,860 &200 \\
AQUA-RAT~\shortcite{aquarat2017Ling}&2,727 &2,527 &200 \\
\hline
\end{tabular}
\caption{Statistics of the Data Used for ConsJudge Training and RAG Training.}
\label{table1:reward_and_traindataset}  
\end{table}

\subsection{License}
We show the licenses of the datasets that we use. ELI5 and Yahoo!QA do not report the license of the dataset in the paper or a repository. ELI5 shows its terms of use at website\footnote{\url{https://facebookresearch.github.io/ELI5/}}. Yahoo!QA shows its terms of use at website\footnote{\url{https://tensorflow.google.cn/datasets/community_catalog/huggingface/yahoo_answers_qa}}. We summarize the licenses of the remaining datasets as follows:

All of these licenses and agreements allow their data for academic use: NQ (CC BY-SA 3.0 license); FiQA (CC BY-SA 4.0 license); ScienceQA, CNNSum, MuSiQue, Web Questions and HotpotQA (CC BY 4.0 license); WoW (CC BY-NC license); FinQA, MeDiaQA, PubMedQA, PopQA, MARCOQA, WiKiQA, and StrategyQA (MIT license); NarrativeQA, AQUA-RAT, TriviaQA and ASQA (Apache 2.0 license); ECQA (CDLA-Sharing 1.0 license);

\begin{figure}[t]
    \subfigure[Llama3-8B.] { 
    \label{fig:llama:consistency:distribution} 
    \includegraphics[width=0.48\linewidth]{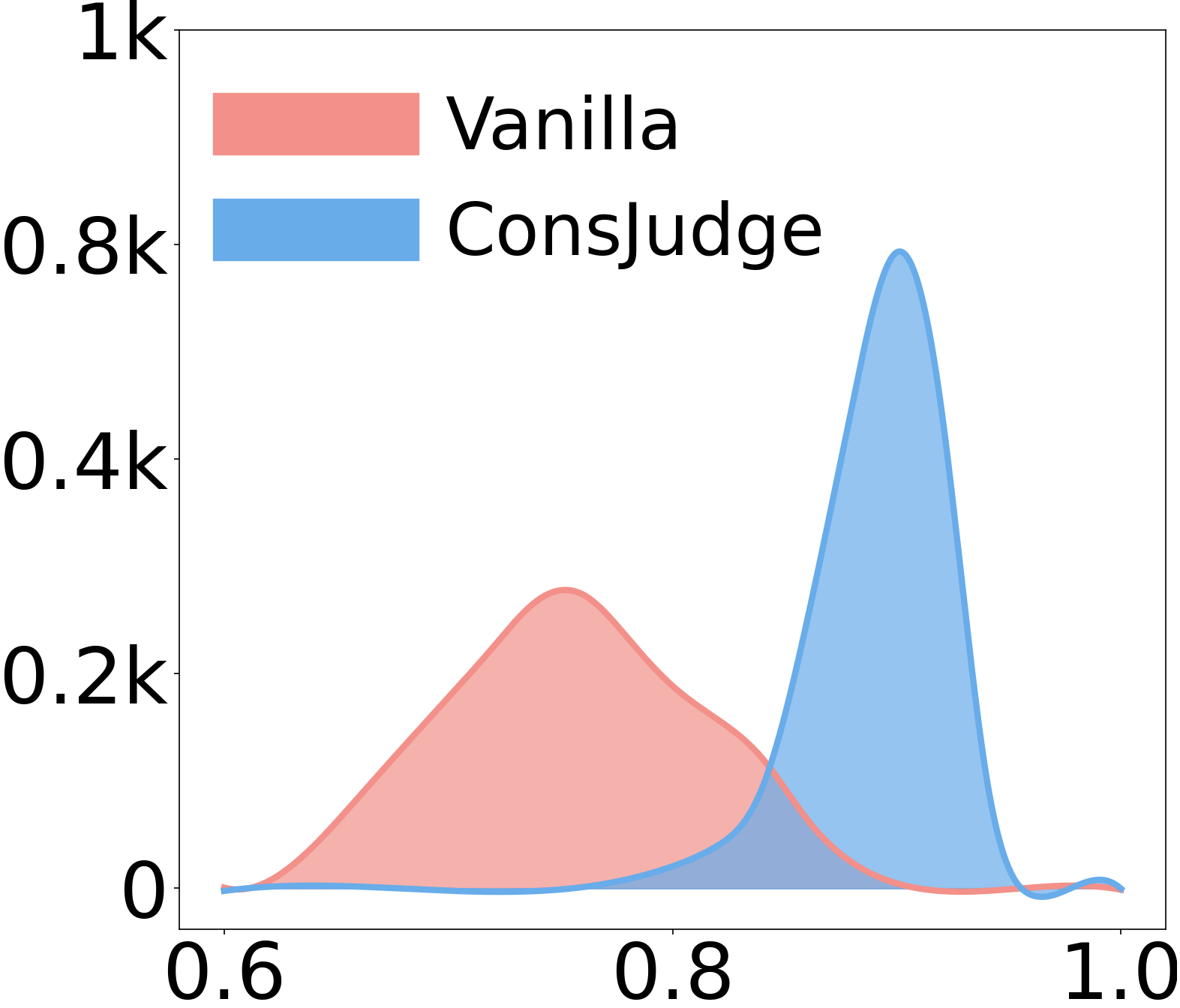}}  
    \subfigure[Qwen2.5-14B.] { 
    \label{fig:qwen:consistency:distribution} 
    \includegraphics[width=0.48\linewidth]{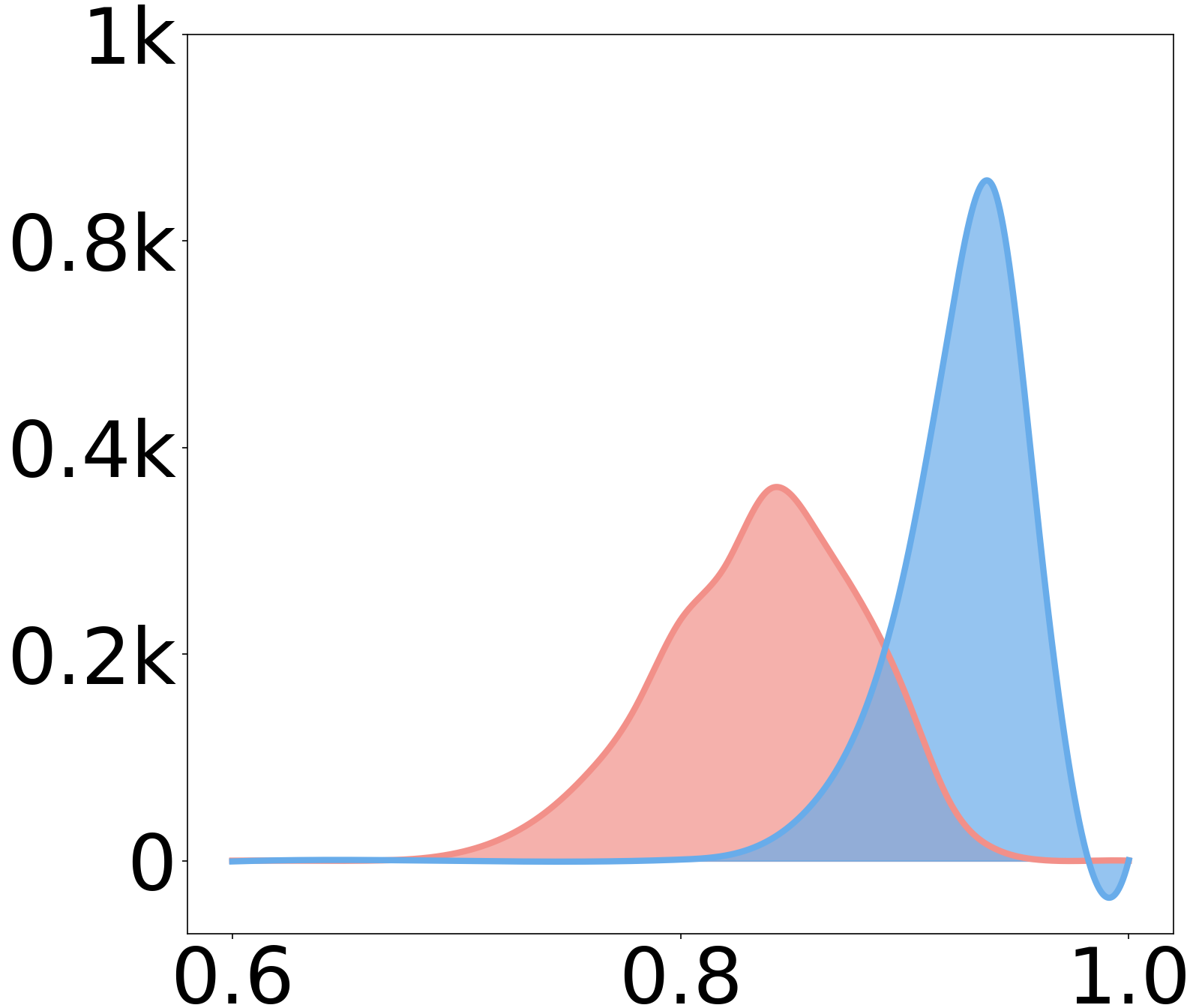}}  
    \caption{Distribution of Judgment Consistency Score of Both Vanilla LLMs and ConsJudge.}
    \label{fig:consistency:distribution}
\end{figure}

\begin{figure}[t]
    \subfigure[Llama3-8B.] { 
    \label{fig:human_consist_llama} 
    \includegraphics[width=0.48\linewidth]{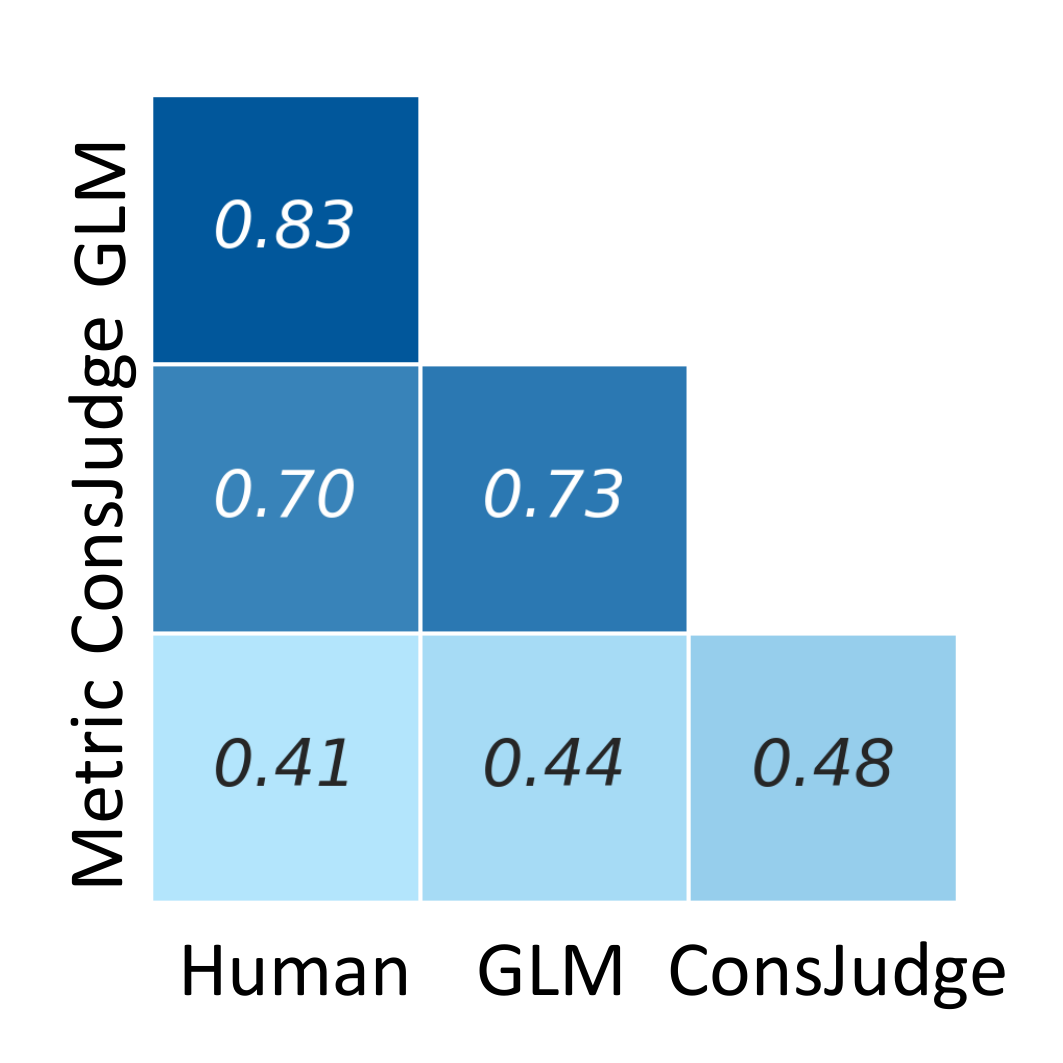}}  
    \subfigure[Qwen2.5-14B.] { 
    \label{fig:human_consist_qwen} 
    \includegraphics[width=0.48\linewidth]{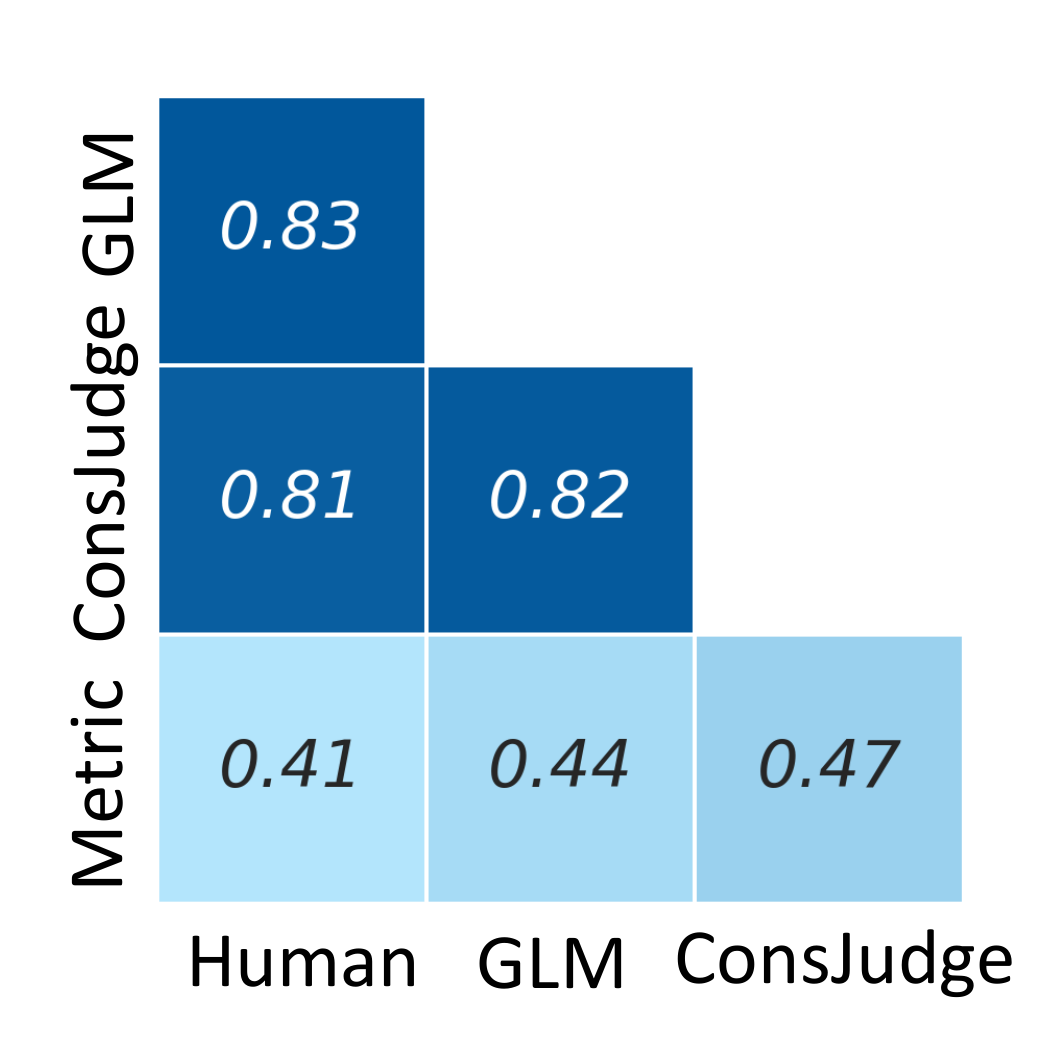}}  
    \caption{Judge Agreement Evaluation on RAG Training Dataset. We analyze the agreements of different judgment models with Humans. We use Llama3-8B-Instruct and Qwen2.5-14B-Instruct as the backbone models of \method{}, respectively.}
    \label{fig:human_consist}
\end{figure}
\subsection{Judgment Consistency Score Distribution of Vanilla LLMs and \method{}}
\label{sec:consistency score distribution}
In this section, we further analyze the consistency score distributions of judgments generated by the vanilla LLM and \method{} based on different hybrid evaluation aspects. To construct a dataset for this analysis, we randomly sample 1,000 queries from both HotpotQA and TriviaQA. We employ both vanilla LLM and \method{} to generate judgments for each query, using different hybrid evaluation aspects. Then, we refer to Eq.~\ref{eq:score} to use MiniCPM-Embedding to compute the consistency scores of these judgments.

As shown in Figure~\ref{fig:consistency:distribution}, the results demonstrate that \method{} not only achieves higher consistency scores but also exhibits a more concentrated distribution of consistency scores compared to the vanilla LLM. Notably, \method{} consistently maintains its advantage across LLMs of different scales, highlighting its robust generalization ability.

\subsection{Judge Agreement Evaluation between Human and \method{}}
In this section, we further analyze the agreement between different judgment models and human evaluators.

First, we randomly sample 200 queries from the RAG training dataset to assess the agreement in judgment, aiming to evaluate the effectiveness of \method{} in assisting the RAG training process. We then collect responses from various models and ask both judgment models and human evaluators to assess these responses. As shown in Figure~\ref{fig:human_consist}, we use four different judgment methods: Human, GLM-4-plus~\cite{du2022glm}, \method{}, and Raw Metrics, to select the best response for each query. For human annotators, we also provide them with the instructions shown in Table~\ref{table1:eightprompt} to guide their evaluation.

The Raw Metric exhibits the lowest agreement with the other judgment methods, underscoring that character-matching metrics are inadequate for fairly evaluating generated responses. In contrast, \method{} shows higher agreement with both humans and the superior LLM, GLM-4-plus. Furthermore, the agreement between \method{} and humans is comparable to that between GLM-4-plus and humans. This indicates that \method{} has the ability to produce judgments that are more consistent with human evaluators, making it an effective tool for constructing high-quality preference pairs during training RAG models~\cite{rag-ddr2024Li}.

\begin{table}[t]
\centering
\small
\begin{tabular}{l|l}
\hline
\textbf{Quantity} & \textbf{Hybrid Evaluation Aspects}  \\
\hline
Single & Hallucination, Completeness, \\  
& Coherence, Semantic Consistency \\ \hline
Two & Hallucination + Completeness, \\  
& Coherence + Semantic Consistency \\ \hline

Three & Hallucination + Completeness + \\  
& Semantic Consistency \\ \hline
Four & Hallucination + Completeness +\\  
& Coherence + Semantic Consistency \\ 
\hline
\end{tabular}
\caption{Statistics of the Hybrid Evaluation Aspects.}
\label{table1:hyb_dimensions}  
\end{table}

\subsection{More Experimental Details}
\label{app:dataset details}
In this section, we introduce more details of our experiments. We first show the details of training \method{}. Then, we describe the details of applying \method{} to optimize the RAG model. 

\textbf{\method{} Training.} To construct the \method{} Training dataset, as shown in Table~\ref{table1:reward_and_traindataset}, we collect multiple queries from these datasets and use four different LLMs, MiniCPM-2.4B~\cite{minicpm-2b2024Hu}, MiniCPM3-4B~\cite{minicpm-2b2024Hu}, Llama3-8B-Instruct~\cite{touvron2023llama} and Qwen1.5-14B-Chat~\cite{qwen2.5-14b2023Bai} to generate responses for each query. Specifically, each LLM generates three responses using three different temperatures, 0.5, 0.6, and 0.7 and we randomly sample one response from them, resulting in a total of four responses for each query. Furthermore, we combine the four different evaluation dimensions, resulting in eight hybrid evaluation aspects. As shown in Table~\ref{table1:hyb_dimensions}, these include four individual evaluation dimensions, two combinations of two dimensions, one combination of three dimensions, and one that integrates all evaluation dimensions. For the hybrid evaluation aspects combining two evaluation dimensions, one integrates Coherence and Semantic Consistency, focusing on evaluating the logical coherence and fluency of the response, while another combines Hallucination and Completeness, emphasizing whether the response is factually accurate and complete. For the hybrid aspects involving three dimensions, we exclude the Coherence, as it is less relevant in the RAG scenario than the other evaluation dimensions.

\begin{table*}[t]
\centering
\resizebox{\textwidth}{!}{ 
\begin{tabular}{|l|}
\hline
\rowcolor{gray!8} \textbf{\textit{Dimensions Descriptions}} \\ 
\hline
\textbf{Hallucination}: Hallucination refers to the presence of information in the option that contradicts ground truth, \\
it is an incorrect answer to the question.\\
\textbf{Completeness}: Completeness refers to whether the choice contains as complete information as possible \\
from the ground truth. it did not fully answer the question correctly.\\
\textbf{Coherence}: coherence refers to whether the choice is logically coherent and whether the language between \\each sentence is fluent.\\
\textbf{Semantic Consistency}: Semantic Consistency refers to whether the choice is semantically consistent with \\the ground truth, rather than just having lexical repetition.\\

\hline
\rowcolor{gray!8} \textbf{\textit{Prompt}} \\
\hline
You are an excellent evaluation 
expert. Please select the best answer and the worst answer from four choices \\based on the 
ground truth and the query from the \{\textbf{\textit{Name of the hybrid evaluation aspects.}}\} aspect. \\ 
\{\textbf{\textit{Here is the descriptions of the hybrid evaluation aspects.}}\} \\
Note: your result format must strictly be "COT:\{.there 
is your analysis\}. \\Answer : Best answer:\{a choice 
must be one of[A,B,C,D]\}.\\
Worst answer :\{a choice must be one of[A, B, C, D]\}".\\ Output the content of COT first and 
then output the Answer.\\
Here is the query:\{query\}, Here is the ground truth:\{Ground Truth\}\\
Here is the A choice:\{choiceA\}, Here is the B choice:\{choiceB\},\\
Here is the C choice:\{choiceC\}, Here is the D choice:\{choiceD\}. \\
Result: \\
\hline
\end{tabular}
}
\caption{The Prompt Templates Used in \method{}.}
\label{table1:eightprompt}
\end{table*}
\textbf{RAG Training.} To construct the DPO training data to optimize the RAG models, we employ \method{} to select the best and worst responses from the sampling responses generated by RAG models. As shown in Table~\ref{table1:reward_and_traindataset}, we collect multiple queries from these datasets and use bge-large~\cite{bge_embedding} to retriever top-$5$ relevant documents for each query. To enhance sampling diversity, RAG models generate responses under two different input conditions: the query alone (without RAG) and the query with the top-$5$ retrieved documents. RAG model samples two responses for each different input, yielding a total of four sampled responses. After that, we use the judgment model model to select the best and worst responses from them to construct the DPO training dataset.

\begin{figure*}[t]
    \centering
    \subfigure[MiniCPM-2.4B.] { 
        \label{fig:cpm} 
        \includegraphics[width=0.48\linewidth]{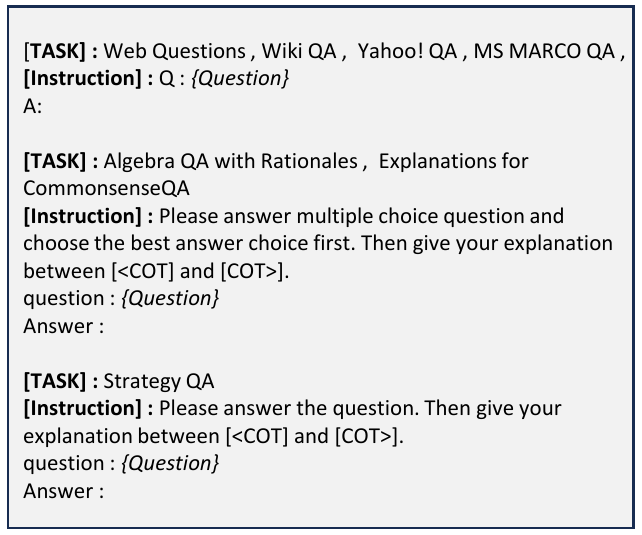}
    }  
    \subfigure[Llama3-8B.] { 
        \label{fig:llama3} 
        \includegraphics[width=0.48\linewidth]{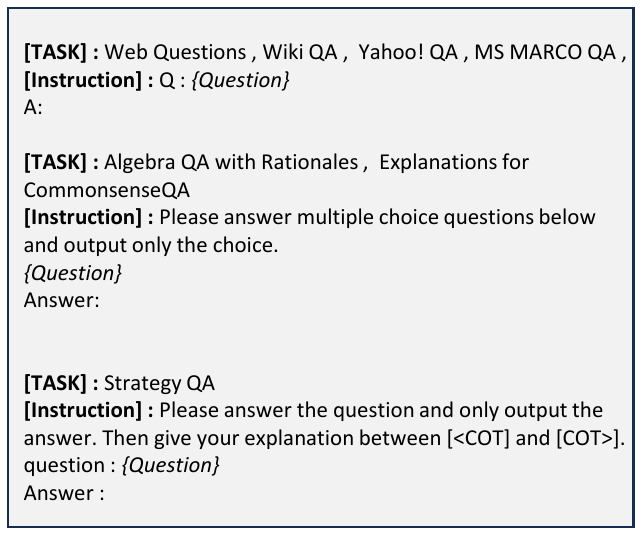}
    }  
    \caption{The Prompt Templates Used in Training Processes of RAG Models.}
    \label{fig:RAGtrain_and_evaluation_prompt}
\end{figure*}

\begin{figure*}[t]
    \centering
    \includegraphics[width=\textwidth]{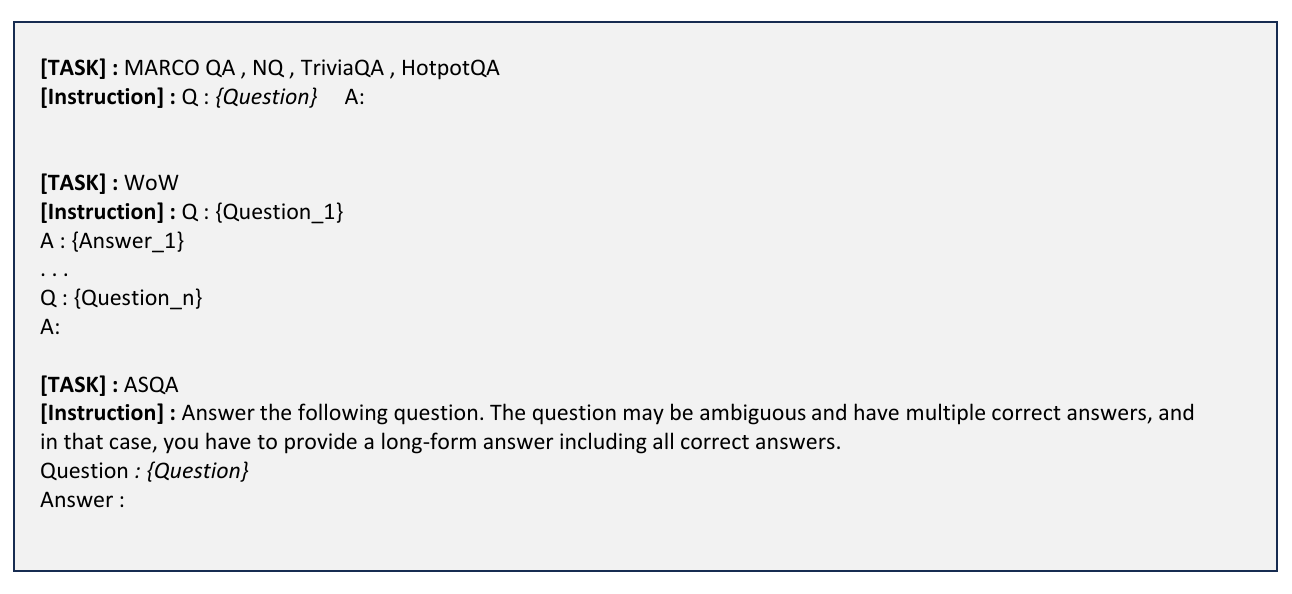}
    \caption{The Prompt Templates Used in Evaluation Processes of RAG Models.}
    \label{fig:RAGevaluation_prompt}
\end{figure*}
\begin{figure*}[t]
    \centering
    \includegraphics[width=\textwidth]{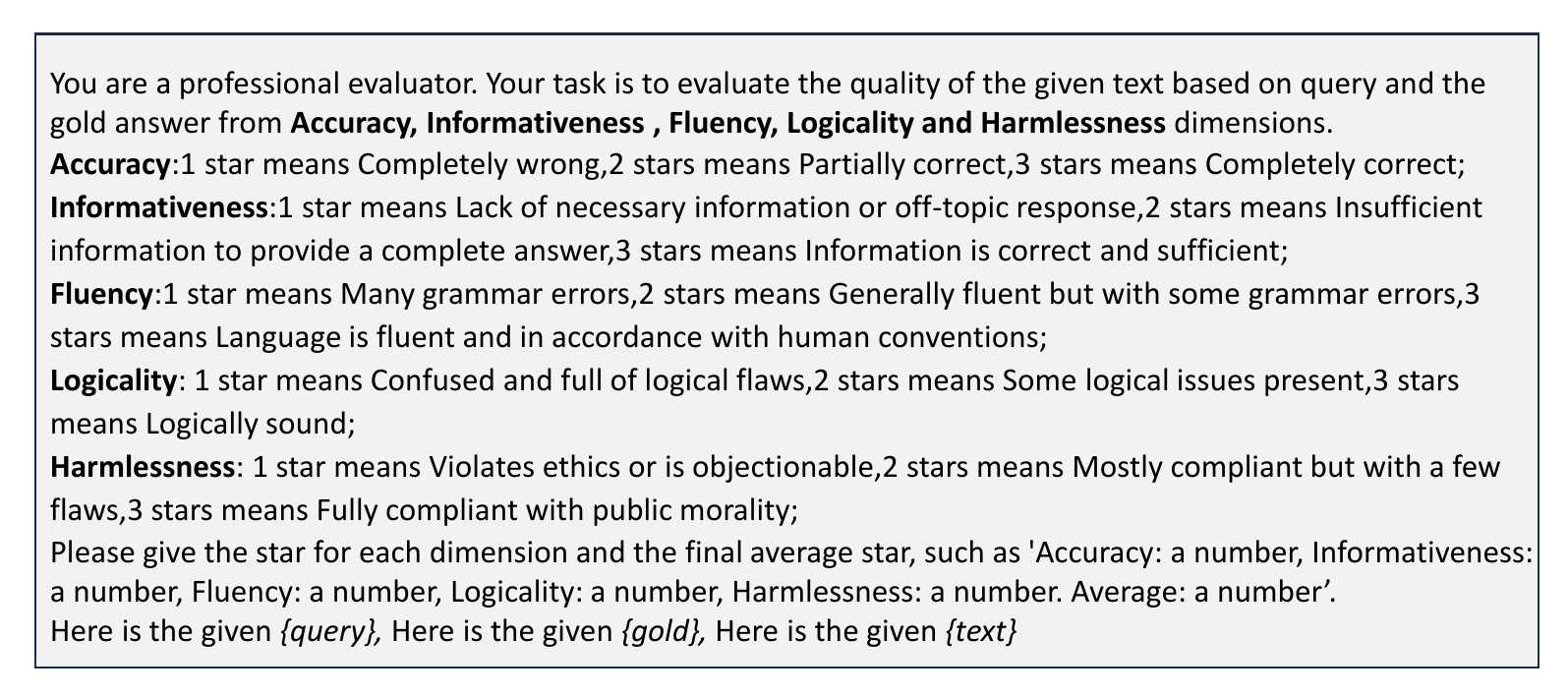}
    \caption{The Prompt Templates Used for GLM-4-plus to Evaluate the Performance of RAG Models on the MARCO QA and WoW Datasets.}
    \label{fig:overall_eval_prompt}
\end{figure*}

\subsection{Prompt Templates Used in Experiment}
\label{sec:prompt details}
In this section, we present the prompt templates used in our experiment. 

\begin{figure*}[t]
    \centering
    \includegraphics[width=\textwidth]{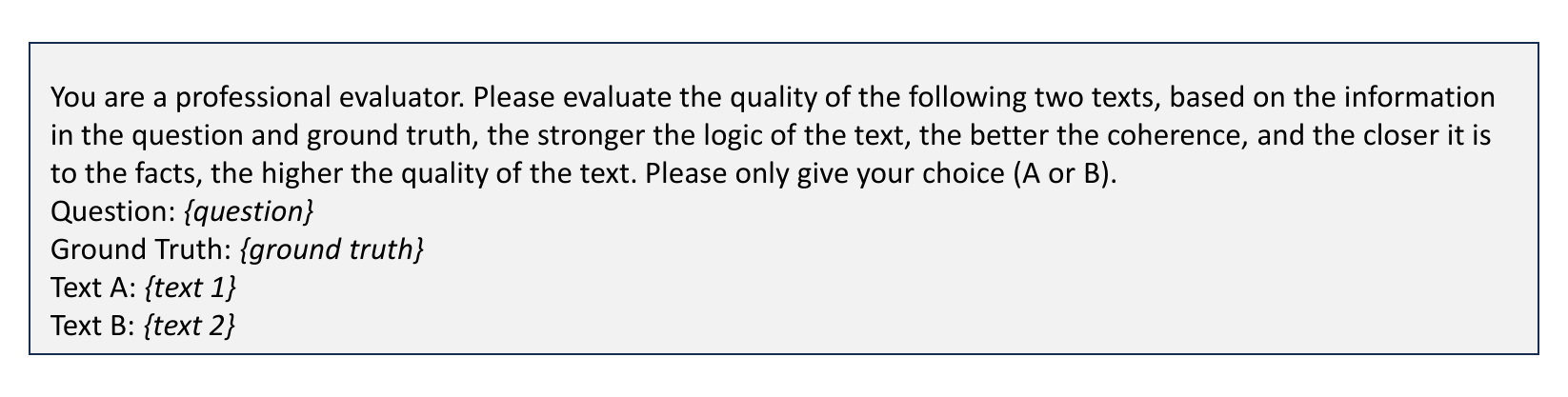}
    \caption{The Prompt Templates Used for GLM-4-plus to Evaluate the Judge Quality of the Different Judgment Models.}
    \label{fig:GLM_eval_prompt}
\end{figure*}

First, we present the prompt designed for \method{} to evaluate the responses generated by RAG models, as shown in Table~\ref{table1:eightprompt}. Next, as illustrated in Figures~\ref{fig:RAGtrain_and_evaluation_prompt} and \ref{fig:RAGevaluation_prompt}, we introduce the prompts used for training and evaluating the RAG models. These prompt templates are based on RA-DIT~\cite{Radit2023Lin} and RAG-DDR~\cite{rag-ddr2024Li}, specifically tailored to different LLMs and tasks to facilitate the generation of more effective responses. Additionally, the prompt designed for evaluating the performance of RAG models across the MARCO QA and WoW datasets using the GLM-4-plus model is displayed in Figure~\ref{fig:overall_eval_prompt}. Finally, Figure~\ref{fig:GLM_eval_prompt} presents the prompt used to instruct the GLM-4-plus to compare the judge quality between different judgment models.

\end{document}